\def\year{2022}\relax
\documentclass[letterpaper]{article} % DO NOT CHANGE THIS
\usepackage{aaai22}  % DO NOT CHANGE THIS
\usepackage{times}  % DO NOT CHANGE THIS
\usepackage{helvet}  % DO NOT CHANGE THIS
\usepackage{courier}  % DO NOT CHANGE THIS
\usepackage[hyphens]{url}  % DO NOT CHANGE THIS
\usepackage{graphicx} % DO NOT CHANGE THIS
\usepackage{natbib}  % DO NOT CHANGE THIS AND DO NOT ADD ANY OPTIONS TO IT
\usepackage{caption} % DO NOT CHANGE THIS AND DO NOT ADD ANY OPTIONS TO IT
\DeclareCaptionStyle{ruled}{labelfont=normalfont,labelsep=colon,strut=off} % DO NOT CHANGE THIS
\usepackage[utf8]{inputenc} % allow utf-8 input
\usepackage[T1]{fontenc}    % use 8-bit T1 fonts
\usepackage{hyperref}       % hyperlinks
\usepackage{url}            % simple URL typesetting
\usepackage{booktabs}       % professional-quality tables
\usepackage{amsfonts}       % blackboard math symbols
\usepackage{nicefrac}       % compact symbols for 1/2, etc.
\usepackage{microtype}      % microtypography
\usepackage{adjustbox}
\usepackage{bbm}
\usepackage{algorithm}
\usepackage{amsmath}
\usepackage{amsfonts} 
\usepackage{subfigure}
\usepackage{microtype}
\usepackage{booktabs}%table
\usepackage{makecell}%table
\usepackage{diagbox}
\usepackage[noend]{algpseudocode}
\usepackage[dvipsnames]{xcolor}
\usepackage{amssymb}
\usepackage{amsthm}
\usepackage{color}
\usepackage{wrapfig}

\def\bc{\bar{c}}

\title{Adversarial Examples can be Effective Data Augmentation \\for Unsupervised Machine Learning}
\author{
    Chia-Yi Hsu\textsuperscript{\rm 1},
    Pin-Yu Chen\textsuperscript{\rm 2},
    Songtao Lu\textsuperscript{\rm 2},
    Sijia Liu\textsuperscript{\rm 3},
    Chia-Mu Yu\textsuperscript{\rm 1}
}
\affiliations{
    \textsuperscript{\rm 1}National Yang Ming Chiao Tung University\\
    \textsuperscript{\rm 2} IBM Research \\
    \textsuperscript{\rm 3} Michigan State University\\
    \{chiayihsu8315, chiamuyu\}@gmail.com,
    \{pin-yu.chen, songtao\}@ibm.com, 
    liusiji5@msu.edu

}
\begin{document}
\maketitle
\begin{abstract}
Adversarial examples causing evasive predictions are widely used to evaluate and improve the robustness of machine learning models. However, current studies focus on supervised learning tasks, relying on the ground-truth data label, a targeted objective, or supervision from a trained classifier. In this paper, we propose a framework of generating adversarial examples for \textbf{unsupervised} models and demonstrate novel applications to data augmentation. Our framework exploits a mutual information neural estimator as an information-theoretic similarity measure to generate adversarial examples without supervision. We propose a new MinMax algorithm with provable convergence guarantees for efficient generation of unsupervised adversarial examples. Our framework can also be extended to supervised adversarial examples. When using unsupervised adversarial examples as a simple plug-in data augmentation tool for model retraining, significant improvements are consistently observed across different unsupervised tasks and datasets, including data reconstruction, representation learning, and contrastive learning. Our results show novel methods and considerable advantages in studying and improving unsupervised machine learning via adversarial examples. 
\end{abstract}

\section{Introduction}
Adversarial examples are known as prediction-evasive attacks on state-of-the-art machine learning models (e.g., deep neural networks), which are often generated by manipulating native data samples while maintaining high similarity measured by task-specific metrics such as $L_p$-norm bounded perturbations \cite{goodfellow2014explaining,biggio2018wild}. Due to the implications and consequences on mission-critical and security-centric machine learning tasks, adversarial examples are widely used for robustness evaluation of a trained model and for robustness enhancement during training (i.e., adversarial training). 

Despite of a plethora of adversarial attacking algorithms, the design principle of existing methods is primarily for \textit{supervised} learning models --- requiring either the true label or a targeted objective (e.g., a specific class label or a reference sample).
Some recent works have extended to the \textit{semi-supervised} setting, by leveraging supervision from a classifier (trained on labeled data) and using the predicted labels on unlabeled data for generating (semi-supervised) adversarial examples \cite{miyato2018virtual,zhang2019theoretically,stanforth2019labels,carmon2019unlabeled}.
On the other hand, recent advances in unsupervised and few-shot machine learning techniques show that task-invariant representations can be learned and contribute to downstream tasks with limited or even without supervision \cite{ranzato2007unsupervised,zhu2009introduction,zhai2019s4l}, which motivates this study regarding their robustness. Our goal is to provide efficient robustness evaluation and data augmentation techniques for unsupervised (and self-supervised) machine learning models through \textit{unsupervised} adversarial examples (UAEs). Table \ref{tab:UAE_illustration} summarizes the fundamental difference between conventional supervised adversarial examples and our UAEs.
Notably, our UAE
generation is supervision-free because it solely uses an information-theoretic similarity measure and the associated unsupervised learning objective function.
It does not use any supervision such as label information or prediction from other supervised models.    

\begin{table}[t]
%\begin{wraptable}{r}{5.7cm}
%\vspace{-1mm}
\centering
%\vspace{-2mm}
\begin{adjustbox}{max width=0.99\columnwidth}
\begin{tabular}{@{}lllll@{}}
\toprule
\multicolumn{5}{l}{(I) \textit{ Mathematical notation}}                                                                                                                                                                                                                                                                                                                                                                                                                  \\ \midrule
\multicolumn{5}{l}{\begin{tabular}[c]{@{}l@{}}$M^{\text{sup}}/M^{\text{unsup}}$: trained supervised/unsupervised machine learning models\\ $x/x_{\text{adv}}$: original/adversarial data sample\\ $\ell^{\text{sup}}_x/\ell^{\text{unsup}}_x$: supervised/unsupervised loss function in reference to $x$\end{tabular}}                                                                                                                                                         \\ \midrule
\multicolumn{1}{l|}{\begin{tabular}[c]{@{}l@{}}(II) \textit{Supervised tasks}\\ (e.g. classification)\end{tabular}}                                                                                          & \multicolumn{4}{l}{\begin{tabular}[c]{@{}l@{}}(III) \textit{\textcolor{blue}{Unsupervised tasks}} (our proposal)\\ (e.g. data reconstruction, contrastive learning)\end{tabular}}                                                  \\ \midrule
\multicolumn{1}{l|}{\begin{tabular}[c]{@{}l@{}}$x_{\text{adv}}$ is \textcolor{red}{similar} to $x$ but\\ $M^{\text{sup}}(x_{\text{adv}}) \neq  M^{\text{sup}}(x)$\end{tabular}}                      & \multicolumn{4}{l}{\begin{tabular}[c]{@{}l@{}}$x_{\text{adv}}$ is \textcolor{red}{dissimilar} to $x$ but\\ $\ell^{\text{unsup}}_x(x_{\text{adv}}|M^{\text{unsup}}) \leq \ell^{\text{unsup}}_x(x|M^{\text{unsup}})$\end{tabular}} \\ \bottomrule
\end{tabular}
\end{adjustbox}
\vspace{-2mm}
\caption{Illustration of adversarial examples for supervised/unsupervised machine learning tasks. Both settings use a native data sample $x$ as reference.
    For supervised setting, adversarial examples refer to \textit{similar} samples of $x$ causing inconsistent predictions. For unsupervised setting, adversarial examples refer to \textit{dissimilar} samples yielding smaller loss than $x$, relating to generalization errors on low-loss samples.}\label{tab:UAE_illustration}
\vspace{-3mm}
%\end{wraptable}
\end{table}

% \begin{figure}[t]
%     \centering
%     \includegraphics[scale=0.43]{Fig_UAE_illustration.png}
%     \vspace{-4mm}
%     \caption{Illustration of adversarial examples for supervised and unsupervised machine learning tasks. Both settings use a native data sample $x$ as reference.
%     For supervised setting, adversarial examples refer to \text{similar} samples of $x$ causing inconsistent model predictions. For unsupervised setting, adversarial examples refer to \textit{dissimilar} samples yielding smaller loss in reference to $x$, which can be interpreted as generalization errors on low-loss samples.
%      }
%      \label{UAE_illustration}
% \end{figure}

In this paper, we aim to formalize the notion of UAE, establish an efficient framework for UAE generation, and demonstrate the advantage of UAEs for improving a variety of unsupervised machine learning tasks. We summarize our main contributions as follows.
\\
$\bullet$ We propose a new per-sample based mutual information neural estimator (MINE) between a pair of original and modified data samples as an information-theoretic similarity measure and a supervision-free approach for generating UAE. For instance, see UAEs for data reconstruction in Figure \ref{UAEs_svhn} of supplementary material.
%To the best of our knowledge, this work is the first study on UAEs. 
While our primary interest is generating adversarial examples for unsupervised learning models, we also demonstrate that our per-sample MINE can be used to generate adversarial examples for supervised learning models with improved visual quality.
\\  
$\bullet$ We formulate the generation of adversarial examples with MINE as a constrained optimization problem, which applies to both supervised and unsupervised machine learning tasks. We then develop an efficient MinMax optimization algorithm (Algorithm \ref{minmax}) and prove its convergence. We also demonstrate the advantage of our MinMax algorithm over the  conventional penalty-based method.  
\\
$\bullet$ We show a novel application of UAEs as a simple plug-in data augmentation tool for several unsupervised machine learning tasks, including data reconstruction, representation learning, and contrastive learning on image and tabular datasets. Our extensive experimental results show outstanding performance gains (up to 73.5\% performance improvement) by retraining the model with UAEs.

\section{Related Work and Background}
\subsection{Adversarial Attack and Defense}
For supervised adversarial examples, the attack success criterion can be either \textit{untargeted} (i.e. model prediction differs from the true label of the corresponding native data sample) or \textit{targeted} (i.e. model prediction targeting a particular label or a reference sample). In addition, a similarity metric such as 
$L_p$-norm bounded perturbation is often used when generating adversarial examples. The projected gradient descent (PGD) attack \cite{madry2017towards} is a widely used approach to find $L_p$-norm bounded supervised adversarial examples.
Depending on the attack threat model,
the attacks can be divided into white-box \cite{szegedy2013intriguing,carlini2017towards}, black-box \cite{chen2017zoo,brendel2017decision,liu2020min}, and transfer-based \cite{bhagoji2017exploring,papernot2017practical} approaches.

Although a plethora of defenses were proposed, many of them failed to withstand advanced attacks \cite{carlini2017adversarial,athalye2018obfuscated}. Adversarial training \cite{madry2017towards} and its variants aiming to generate worst-case adversarial examples during training are so far the most effective defenses. However, adversarial training on supervised adversarial examples can suffer from undesirable tradeoff between robustness and accuracy \cite{su2018robustness,tsipras2019robustness}. 
Following the formulation of untargeted supervised attacks, recent studies such as \cite{cemgil2019adversarially} generate adversarial examples for unsupervised tasks by finding an adversarial example within an $L_p$-norm perturbation constraint that maximizes the training loss. In contrast, our approach aims to find adversarial examples that have low training loss but are dissimilar to the native data (see Table \ref{tab:UAE_illustration}), which plays a similar role to the category of ``on-manifold'' adversarial examples governing generalization errors \cite{stutz2019disentangling}.
In supervised setting, \cite{stutz2019disentangling} showed that adversarial training with $L_p$-norm constrained perturbations may find
off-manifold adversarial examples and hurt generalization.

%It is worth noting that the scope of current adversarial attacks and defenses mainly focuses on supervised adversarial examples, which in turn excludes machine learning tasks without supervision, such as data reconstruction and representation learning. In this paper, we aim to bridge this gap and demonstrate the utility of UAEs for several unsupervised machine learning tasks.

%\subsection{Difference to Adversarial Examples Generated from Unlabeled Data and Class Predictions}

%image scaling; super-resolution

%Szegedy \emph{et al.} make attacking deep neural networks as an optimization problem \cite{szegedy2013intriguing}. The purpose of their approach is finding a minimal perturbation of image pixel that lets well-trained classifier misclassified. There have been many attacks following the same method but with different formulations. Also using $L_{p}$-norm to restrain perturbations so that adversarial examples are imperceptible for human eyes. Madry \emph{et al.} propose PGD attack which is a gradient-based attack and restricts perturbations with $L_{\infty}$ -norm \cite{madry2017towards}. Carlini and Wanger provided
%\subsection{Semantic attack}
\subsection{Mutual Information Neural Estimator%(MINE)
}

Mutual information (MI) measures the mutual dependence between two random variables $X$ and $Z$, defined as
$I(X,Z) = H(X) - H(X|Z)$,
where $H(X)$ denotes the (Shannon) entropy of $X$ and $H(X|Z)$ denotes the conditional entropy of $X$ given $Z$.
% Although MI is 
 %a widely used information-theoretic similarity measure, 
 %and is 
 %widely used to characterize the  fundamental capacity and limitation of machine learning models, such as the information bottleneck of neural networks \cite{tishby2000information}, 
Computing MI can be difficult without knowing the marginal and joint probability distributions ($\mathbb{P_{X}}$, $\mathbb{P_{Z}}$, and $\mathbb{P_{XZ}}$).
For efficient computation, the mutual information neural estimator (MINE) with consistency guarantees is proposed in \cite{belghazi2018mutual}. Specifically, MINE aims to maximize the lower bound of the exact MI using a model parameterized by a neural network $\theta$, defined as  $I_{\Theta}(X,Z) \leq I(X,Z)$,
 where $\Theta$ is the space of feasible parameters of a neural network, and $I_{\Theta}(X,Z)$ is the neural information quantity defined as $I_{\Theta}(X,Z) = \mathop{{\rm sup}}_{\theta \in \Theta}\mathbb{E}_{\mathbb{P}_{XZ}}[T_{\theta}] - {\rm log}(\mathbb{E}_{\mathbb{P}_{X} \otimes \mathbb{P}_{Z}}[e^{T_{\theta}}])$.
The function $T_\theta$ is parameterized by a neural network $\theta$ based on the Donsker-Varadhan representation theorem \cite{donsker1983asymptotic}. MINE estimates the expectation of the quantities above by shuffling the samples from the joint distribution along the batch axis or using empirical samples $\{x_i,z_i\}_{i=1}^n$ from $\mathbb{P_{XZ}}$ and $\mathbb{P_{X} \otimes P_{Z}}$ (the product of marginals).

MINE has been successfully applied to improve representation learning \cite{hjelm2018learning,zhu2020learning} given a dataset. However, for the purpose of generating an adversarial example for a given data sample, the vanilla MINE is not applicable because it only applies to a batch of data samples (so that empirical data distributions can be used for computing MI estimates) but not to single data sample. To bridge this gap, we will propose two MINE-based sampling methods for single data sample in Section \ref{subsec_MINE_single}.

%to enable the computation of MI via the vanilla MINE, which will be discussed in Section \ref{subsec_MINE_single}. 

\section{Methodology}

\subsection{MINE of Single Data Sample }
\label{subsec_MINE_single}
%1. MINE was proposed to image batches, not single image
%According to Belghazi's algorithm, the input of MINE is image batches.
%The vanilla MINE algorithm \cite{belghazi2018mutual} requires a batch of samples for MI computation, which prevents its use in generating an adversarial example of single data sample, namely, 
Given a data sample $x$ and its perturbed sample $x+\delta$, we construct an auxiliary distribution using their random samples or convolution outputs to compute MI via MINE as a similarity measure, which we denote as ``per-sample MINE''.

%To compute the MI between a given data sample $x$ and its perturbed sample $x+\delta$, denoted as the per-sample MINE $I_{\Theta}(x,x+\delta)$,
%we propose two solutions, \textit{random sampling} and \textit{convolution output}.

%However, we compute mutual information between a pair of images which is not satisfied with conditions of MINE. To fit conditions of MINE, we use random sampling and output of convolutional layer, respectively.

%2. Extend to single image (conv, rand sampling)
\textbf{Random Sampling} Using compressive sampling \cite{candes2008introduction}, we perform independent Gaussian sampling of a given sample $x$ to obtain a batch of $K$ compressed samples $\{x_k,(x+\delta)_k\}_{k=1}^K$ for computing $I_{\Theta}(x,x+\delta)$ via MINE. We refer the readers to the supplementary material (SuppMat \ref{appen_MINE}, \ref{append_K}) for more details. We also note that random sampling is agnostic to the underlying machine learning model since it directly applies to the data sample.

%We replicate a pair of images to several pairs of images and generate various Gaussian matrices with the specific standard deviation. Then, the result of multiplying images by matrices is used to be the input of MINE. 

%\begin{table}
%\begin{minipage}[t]{.45\textwidth}
%    \vspace{-3mm}
%    \centering
%    \includegraphics[width=1\textwidth]{images/con_vs_rand.pdf}
%    \captionof{figure}{Visual comparison of MINE-based supervised adversarial examples (untargeted attack with $\epsilon=1$) on CIFAR-10. %Both random sampling and convolution output can be used to craft adversarial examples with high similarity.
    %We generate batches of images for MINE with different methods. }
 %    \label{random_conv}
     
%\end{minipage}
%\hfill
%\begin{minipage}[t]{.49\textwidth}
%\vspace{-1mm}
 %   \centering
        
%    \begin{adjustbox}{max width=1\columnwidth}
 %   \begin{tabular}{ccc}
 %   \toprule Method
  %   & FID &KID\\
 %    \midrule
 %       \makecell[c]{Random Sampling \\(10 runs, $K=96$)} & %$339.47 \pm 8.07$& $14.86 \pm 1.45$\\
  %%     \makecell[c] {1st Convolution\\ Layer Output ($K=96$)} & 344.231 & 10.78\\
  %  \bottomrule 
  %  \label{fid_vs_kid}
  %  \end{tabular}
   % \end{adjustbox}
%    \caption{Frechet and kernel inception distances (FID/KID) between the untargeted adversarial examples of 1000 test samples and the training data in 
%    CIFAR-10  for the proposed per-sample MINEs. }
%\end{minipage}
%\vspace{-1mm}
%\end{table}

\textbf{Convolution Layer Output} 
When the underlying neural network model uses a convolution layer to process the input data (which is an almost granted setting for image data), we propose to use the output of the first convolution layer of a data input, denoted by $conv(\cdot)$, to obtain  $K$ feature maps $\{conv(x)_k,conv(x+\delta)_k\}_{k=1}^K$ for computing $I_{\Theta}(x,x+\delta)$.
We provide the detailed algorithm for convolution-based per-sample MINE in SuppMat \ref{appen_MINE}. 

\begin{figure}[t]
    \centering
    \includegraphics[width=0.45\textwidth]{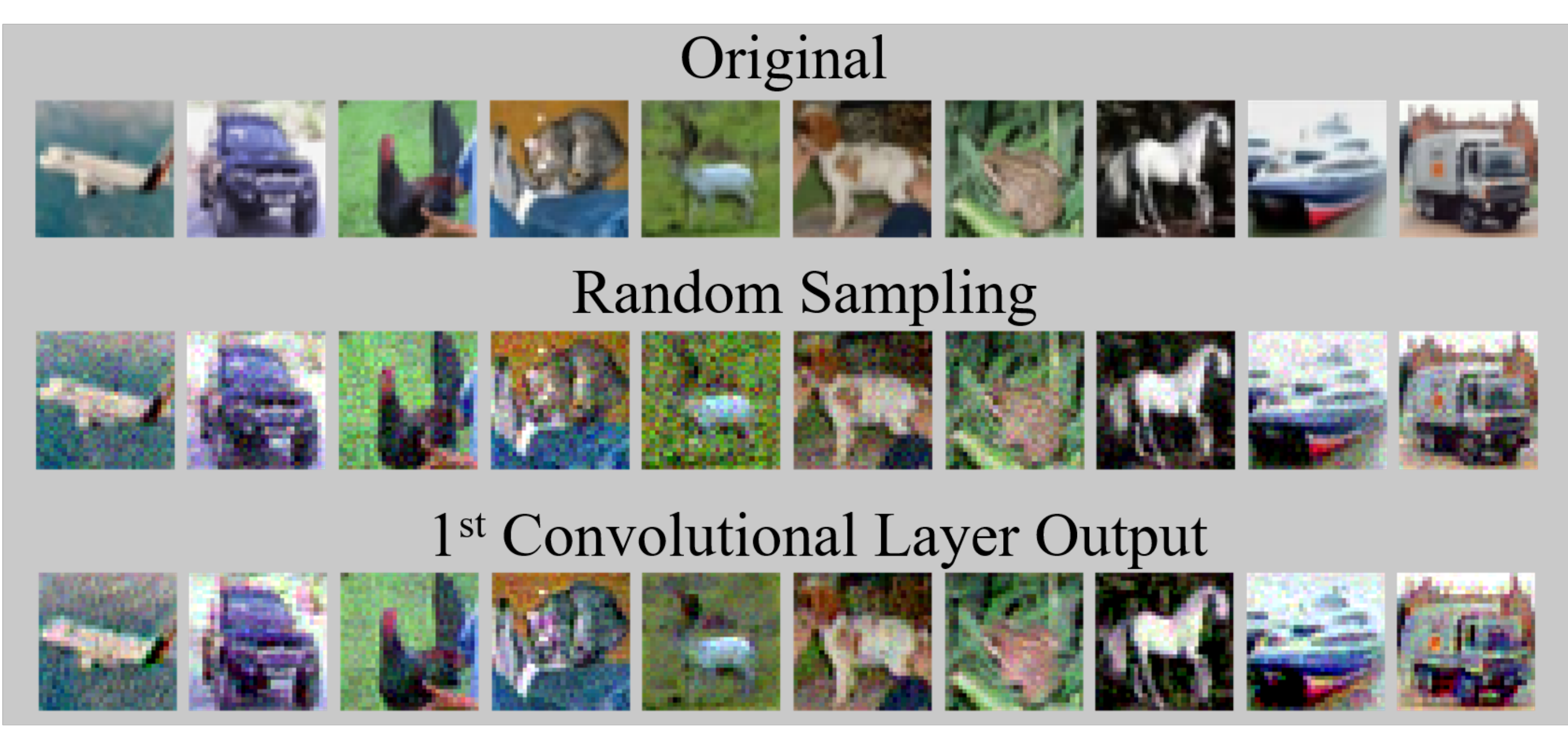}
     \vspace{-3mm}
    \caption{Visual comparison of MINE-based untargeted supervised adversarial examples (with $\epsilon=1$) on CIFAR-10.}
     \label{random_conv}
     \vspace{-2mm}
\end{figure}

\textbf{Evaluation}
We use the CIFAR-10 dataset and the same neural network as in Section \ref{subsec_penality} to provide qualitative and quantitative evaluations on the two per-sample MINE methods for image classification.  
Figure \ref{random_conv} shows their visual comparisons, with the objective of finding the most similar perturbed sample (measured by MINE with the maximal scaled $L_\infty$ perturbation bound $\epsilon=1$) leading to misclassification. Both random sampling and convolution-based approaches can generate high-similarity prediction-evasive adversarial examples despite of large $L_\infty$ perturbation. 

\begin{table}[t]
    \begin{adjustbox}{max width=0.99\columnwidth}
    \begin{tabular}{ccc}
    \toprule Per-sample MINE Method
     & FID &KID\\
     \midrule
        \makecell[c]{Random Sampling \\(10 runs, $K=96$)} & $339.47 \pm 8.07$& $14.86 \pm 1.45$\\
          \midrule
        \makecell[c] {1st Convolution\\ Layer Output ($K=96$)} & 344.231 & 10.78\\
    \bottomrule 
    \end{tabular}
    \end{adjustbox}
    \vspace{-2mm}
    \caption{Frechet and kernel inception distances (FID/KID) between the untargeted adversarial examples of 1000 test samples and the training data in 
    CIFAR-10. }
    \label{fid_vs_kid}
    \vspace{-4mm}
\end{table}

Table \ref{fid_vs_kid} compares the Frechet inception distance (FID) \cite{heusel2017gans} and the kernel inception distance (KID) 
\cite{binkowski2018demystifying} 
between the generated adversarial examples versus the training data (lower value is better). Both per-sample MINE methods have comparable scores. The convolution-based approach attains lower KID score and is observed to have better visual quality as shown in Figure \ref{random_conv}.
We also tested the performance using the second convolution layer output but found degraded performance. In this paper we use convolution-based approach whenever applicable and otherwise use random sampling. 
%For random sampling, we run 10 times and compute average FID and KID scores. Both methods get close results of FID and KID shown in 

%Fig: conv vs rand sampling on FID and KID

\subsection{MINE-based Attack Formulation}
\label{subsec_MINE_attack_formulation}
We formalize the objectives for supervised/unsupervised adversarial examples using per-sample MINE. As summarized in Table \ref{tab:UAE_illustration}, the supervised setting aims to find \textit{most similar} examples causing prediction evasion, leading to an MINE \textit{maximization} problem. The unsupervised setting aims to find \textit{least similar} examples but having smaller training loss, leading to an MINE \textit{minimization} problem. Both problems can be solved efficiently using our unified MinMax algorithm.
%For the sake of brevity, in the remaining paper we use the term `MINE' to mean per-sample MINE.

%\SL{[needs some motivation on why promoting MI in attack generation?
%]}

% \begin{figure}[t]
%     \centering
%     \includegraphics[scale=0.28]{images/con_vs_rand.pdf}
%     \vspace{-4mm}
%     \caption{Visual comparison of MINE-based supervised adversarial examples (untargeted attack with $\epsilon=1$) on CIFAR-10. Both random sampling and convolution output can be used to craft adversarial examples with high similarity.
%     %We generate batches of images for MINE with different methods. 
%      }
%      \label{random_conv}
%      \vspace{-4mm}
% \end{figure}

% \begin{table}[t]
%     \centering
%         \caption{Frechet and kernel inception distances (FID/KID) between the untargeted adversarial examples of 1000 test samples and the training data in 
%     CIFAR-10  for the proposed per-sample MINEs. }
%     \begin{adjustbox}{max width=1\columnwidth}
%     \begin{tabular}{ccc}
%     \toprule 
%      & \makecell[c]{Random Sampling (10 runs, $K=96$)} &\makecell[c] {1st Convolution Layer Output ($K=96$)}\\
%      \midrule
%         FID & $339.47 \pm 8.07$& 344.231\\
%         KID & $14.86 \pm 1.45$ & 10.78\\
%     \bottomrule 
%     \label{fid_vs_kid}
%     \end{tabular}
%     \end{adjustbox}
%     \vspace{-6mm}
% \end{table}

\textbf{Supervised Adversarial Example}
%1. Original formulation: Max MI, such that f=0 (attack is successful)
%We introduce mutual information in supervised attack and design loss function intuitively as following:
Let $(x,y)$ denote a pair of a data sample $x$ and its ground-truth label $y$. The objective of supervised adversarial example is to find a perturbation $\delta$ to $x$ such that the MI estimate $I_{\Theta}(x,x+\delta)$ is maximized while the
prediction of $x+\delta$ is different from $y$ (or being a targeted class $y^\prime \neq y$), which is formulated as 
\[\mathop{{\rm Maximize}}\limits_{\delta}\quad  I_{\Theta}(x,x+\delta )\] 
\[{\rm \textit{such}\,\textit{that}} \; x +\delta \in [0,1]^{d} \; {,} \; \delta \in [-\epsilon,\epsilon]^{d} \; {\rm and} \; f_x(x + \delta) \leq 0. \]
The constraint $x +\delta \in [0,1]^{d}$ ensures $x+\delta$ lies in the (normalized) data space of dimension $d$, and the constraint $\delta \in [-\epsilon,\epsilon]^{d}$ corresponds to the typical bounded $L_\infty$ perturbation norm. We include this bounded-norm constraint to make direct comparisons to other norm-bounded attacks. One can ignore this constraint by setting $\epsilon=1$.
Finally, the function $f^{\text{sup}}_x(x + \delta)$ is an attack success evaluation function, where  $f^{\text{sup}}_x(x + \delta) \leq 0$ means $x+\delta$ is a prediction-evasive adversarial example. For untargeted attack one can use the attack function $f^{\text{sup}}_x$ designed in \cite{carlini2017towards}, which is 
%\[
$f^{\text{sup}}_x(x^{\prime}) = {\text{logit} }(x^{\prime})_{y} - \max_{j: j \neq  y} {\text{logit}} (x^{\prime})_{j} + \kappa$,
%\]
where $\text{logit}(x^{\prime})_j$ is the $j$-th class output of the logit (pre-softmax) layer of a neural network, and $\kappa \geq 0$ is a tunable gap between the original prediction $\text{logit}(x^{\prime})_{y}$ and the top prediction $\max_{j: j \neq  y} \text{logit} (x^{\prime})_{j}$ of all classes other than $y$.  
Similarly, the attack function for targeted attack with a class label $y^\prime \neq y$ is
%\begin{align}
    $f^{\text{sup}}_x(x^{\prime}) =  
\max_{j: j \neq  y^\prime} {\text{logit}} (x^{\prime})_{j} -
{\text{logit}}(x^{\prime})_{y^\prime}  + \kappa $.
%\nonumber
%\end{align}

\textbf{Unsupervised Adversarial Example} Many machine learning tasks such as data reconstruction and unsupervised representation learning do not use data labels, which prevents the use of aforementioned supervised attack functions. Here we use an autoencoder $\Phi(\cdot)$ for data reconstruction to illustrate the  unsupervised attack formulation. The design principle can naturally extend to other unsupervised tasks. The autoencoder $\Phi$ takes a data sample $x$ as an input and outputs a reconstructed data sample $\Phi(x)$. Different from the rationale of supervised attack, for unsupervised attack we propose to use MINE to find the \emph{least similar} perturbed data sample $x+\delta$ with respect to $x$ while ensuring the reconstruction loss of $\Phi(x+\delta)$ is no greater than $\Phi(x)$ (i.e., the criterion of successful attack for data reconstruction). The unsupervised attack formulation is as follows:
\[\mathop{{\rm Minimize}}\limits_{\delta}\quad  I_{\Theta}(x,x+\delta )\]
\[{\rm \textit{such}\,\textit{that}} \; x +\delta \in [0,1]^{d} \; {,} \; \delta \in [-\epsilon,\epsilon]^{d} \; {\rm and} \; f_x(x + \delta) \leq 0 \] 
The first two constraints regulate the feasible data space and the perturbation range. For the $L_2$-norm reconstruction loss, the unsupervised attack function is
\[f^{\text{unsup}}_x(x+\delta) =   \lVert x - \Phi(x+\delta)\rVert_{2} - \lVert x - \Phi(x)\rVert_{2} + \kappa \]
which means the attack is successful (i.e., $f^{\text{unsup}}_x(x+\delta) \leq 0$) if the reconstruction loss of $x+\delta$ relative to the original sample $x$ is smaller than the native reconstruction loss minus a nonnegative margin $\kappa$. That is, $\lVert x - \Phi(x+\delta)\rVert_{2} \leq \lVert x - \Phi(x)\rVert_{2} - \kappa$. In other words, our unsupervised attack formulation aims to find that most dissimilar perturbed sample $x+\delta$ to $x$ measured by MINE while having smaller reconstruction loss (in reference to $x$) than $x$.  Such UAEs thus relates to generalization errors on low-loss samples because the model is biased toward these unseen samples. 
%In Section \ref{subsec_reconst}, we validate the importance of using MINE in unsupervised attack by showing that its performance significantly outruns the alternative objective of maximizing $L_2$ reconstruction loss.

\subsection{MINE-based Attack Algorithm}

Here we propose a unified MinMax algorithm for solving the aforementioned supervised and unsupervised attack formulations, and provide its convergence proof in Section \ref{subsec_minmax_proof}. For simplicity, we will use $f_x$ to denote the attack criterion for $f^{\text{sup}}_x$ or $f^{\text{unsup}}_x$.
%\SL{[ill sentence: minimize or maximize? Check the following two sentences.]}
Without loss of generality, we will analyze the supervised attack objective of maximizing $I_{\Theta}$ with constraints. The analysis also holds for the unsupervised case since minimizing $I_{\Theta}$ is equivalent to maximizing $I'_{\Theta}$, where $I'_{\Theta}=-I_{\Theta}$.
We will also discuss a penalty-based algorithm as a comparative method to our proposed approach.

%   \begin{wrapfigure}{L}{0.5\textwidth}
%     \begin{minipage}{0.5\textwidth}
%       \begin{algorithm}[H]
%         \caption{assignment algorithm}
%         \begin{algorithmic}
%           \STATE i $\leftarrow$ j
%         \end{algorithmic}
%       \end{algorithm}
%     \end{minipage}
%   \end{wrapfigure}

\textbf{MinMax Algorithm (proposed)}
We reformulate the attack generation via MINE as the following MinMax optimization problem with simple convex set constraints:
{\small
\[\mathop{{\mathrm{Min}}}\limits_{\delta: x +\delta \in [0,1]^{d},~\delta \in [-\epsilon,\epsilon]^{d} }\mathop{{\mathrm{Max}}}\limits_{c\geq 0}\quad F(\delta,c) \triangleq c\cdot f_x^+(x+\delta)-I_{\Theta}(x,x+\delta )\]}
The outer minimization problem finds the best perturbation $\delta$ with data and perturbation feasibility constraints $x +\delta \in [0,1]^{d}$ and $\delta \in [-\epsilon,\epsilon]^{d}$, which are both convex sets with known analytical projection functions. The inner maximization associates a variable $c \geq 0$ with the original attack criterion $f_x(x+\delta) \leq 0$, where $c$ is multiplied to the ReLU activation 
function of $f_x$, denoted as $f_x^+(x+\delta)=\text{ReLU}(f_x(x+\delta)) = \max\{f_x(x+\delta),0\}$. The use of $f_x^+$ means when the attack criterion is not met (i.e., $f_x(x+\delta) > 0$), the loss term $c \cdot f_x(x+\delta)$ will appear in the objective function $F$. On the other hand, if the attack criterion is met (i.e., $f_x(x+\delta) \leq  0$), then $c \cdot f_x^+(x+\delta )=0$ and the objective function $F$ only contains the similarity loss term $- I_\Theta(x,x+\delta)$. Therefore, the design of $f_x^+$ balances the tradeoff between the two loss terms associated with attack success and MINE-based similarity.  
We propose to use alternative projected gradient descent between the inner and outer steps to solve the MinMax attack problem, which is summarized in Algorithm \ref{minmax}. The parameters $\alpha$ and $\beta$ denote the step sizes of the minimization and maximization steps, respectively. 
The gradient  $\nabla f_x^+(x+\delta)$  with respect to $\delta$ is set to be 0 when $f_x(x+\delta) \leq 0$.
Our MinMax algorithm returns the successful adversarial example $x+\delta^*$ with the best MINE value $I^*_{\Theta}(x,x+\delta^*)$ over $T$ iterations.
\begin{algorithm}[h]
\caption{MinMax Attack Algorithm}\label{minmax}
\begin{algorithmic}[1]
\State  {\bfseries Require:} data sample $x$, attack criterion $f_x(\cdot)$, step sizes $\alpha$ and $\beta$, perturbation bound $\epsilon$, \# of iterations $T$
\State  Initialize $\delta_0=0$, $c_0=0$, $\delta^*=\text{null}$, $I^*_{\Theta}=-\infty$, $t=1$
\For{$t$ in $T$ iterations}
\State  $\delta_{t+1}$ = $\delta_{t} - \alpha \cdot (c \cdot \nabla f_x^+(x+\delta_{t}) - \nabla I_{\Theta}(x,x+\delta_{t})) $
\State  Project $\delta_{t+1}$ to $[-\epsilon, \epsilon]$ via clipping
\State  Project $x+\delta_{t+1}$ to $[0,1]$ via clipping
\State  Compute $I_{\Theta}(x, x + \delta_{t+1})$
\State  Perform $c_{t+1}= (1 - \frac{\beta}{t^{1/4}})\cdot c_t + \beta \cdot f_x^+(x+\delta_{t+1})$
%\State Perform $c_{t+1}= (1 - \beta\gamma_t)\cdot c_t + \beta \cdot
%f_x^+(x+\delta_{t+1})$
\State  Project $c_{t+1}$ to $[0, \infty]$
\If { $f_x(x+\delta_{t+1}) \leq 0$ and $I_{\Theta}(x, x+\delta_{t+1})> I^*_{\Theta}$ }
\State  update $\delta^* =  \delta_{t+1}$ and $I^*_{\Theta} = I_{\Theta}(x, x+\delta_{t+1})$
\EndIf
\EndFor
\State   {\bfseries Return} $\delta^*$, $I^*_{\Theta}$
\end{algorithmic}
\end{algorithm}

\textbf{Penalty-based Algorithm (baseline)}
%${\rm Z}(x^{\prime})$ is the output of logit layer (pre-softmax %layer) and $l_{x}$ is the ground truth label.$\kappa$ is the confidence of misclassification occurring. Our attack is successful when loss $f =0$ and can find small perturbations by maximizing MI. \\
%2. normally, convert to min c*f - MI - CW use binary search on c to find best delta given f=0
%Normally, we convert the objection function as the form of C\&W attack but substitute negative MI for $L_{2}$-norm regularized loss. 
An alternative approach to solving the MINE-based attack formulation is the penalty-based method with the objective:
\[\mathop{{\rm Minimize}}\limits_{\delta: x +\delta \in [0,1]^{d},~\delta \in [-\epsilon,\epsilon]^{d}}\quad c \cdot f_x^+(x + \delta)- I_{\Theta}(x,x+\delta )\] 
where $c$ is a fixed regularization coefficient instead of an optimization variable. Prior arts such as \cite{carlini2017towards} use a binary search strategy for tuning $c$ and report the best attack results among a set of $c$ values. In contrast, our MinMax attack algorithm dynamically adjusts the $c$ value in the inner maximization stage (step 8 in Algorithm \ref{minmax}). In Section \ref{subsec_penality}, we will show that our MinMax algorithm is more efficient in finding MINE-based adversarial examples than the penalty-based algorithm. The details of the binary search process are given in SuppMat \ref{appen_binary}. Both methods have similar computation complexity involving $T$ iterations of gradient and MINE computations.

\subsection{Convergence Proof of MinMax Attack}
\label{subsec_minmax_proof}

As a theoretical justification of our proposed MinMax attack algorithm (Algorithm \ref{minmax}), we provide a convergence proof with the following assumptions on the considered problem:
\\
$\bullet$ \textbf{A.1:} The feasible set $\Delta$ for $\delta$ is compact, and  $f_x^+(x+\delta)$ has (well-defined) gradients and Lipschitz continuity (with respect to $\delta$) with constants $L_f$ and $l_f$. That is, $|f_x^+(x+\delta)-f_x^+(x+\delta')|\le l_f\|\delta-\delta'\|$ and $\|\nabla f_x^+(x+\delta))-\nabla f_x^+(x+\delta')\|\le L_f\|\delta-\delta'\|,$ ~$\forall~\delta,\delta' \in \Delta$. Moreover,
$I_{\Theta}(x,x+\delta)$ also has gradient Lipschitz continuity with constant $L_I$.
\\
$\bullet$  \textbf{A.2:} %Further, we assume that the mutual information estimator parametrized by the neural nets  can approximate the
The per-sample MINE is $\eta$-stable over iterations for the same input, 
 $|I_{\Theta_{t+1}}(x,x+\delta_{t+1})-I_{\Theta_t}(x,x+\delta_{t+1})|\le\eta$.

%$\bullet$ \textbf{A.3:} The data dimension $d$ is finite and the feasible set of $\delta$ is compact. Moreover, $f_x^+(x+\delta)$ is upper bounded by $F'$ and lower bounded by 0. \PY{???}

A.1 holds in general for neural networks since the numerical gradient of ReLU activation can be efficiently computed and the sensitivity (Lipschitz constant) against the input perturbation can be bounded \cite{weng2018evaluating}. The feasible perturbation set $\Delta$ is compact when the data space is bounded.
A.2 holds by following the consistent estimation proof of the native MINE in \cite{belghazi2018mutual}.
%A.3 also holds in general when that the range of the data values is bounded.

%\subsection{Convergence Results}
To state our main theoretical result, we first define the proximal gradient of the objective function as
$\mathcal{L}(\delta,c):=\left[\delta-P_{\Delta}[\delta-\nabla_{\delta} F(\delta,c)],
c-P_{\mathcal{C}}[c+\nabla_{c} F(\delta,c)] \right]  $,
% \begin{equation}
% \label{eq.optgap}
% { \small
% \mathcal{L}(\delta,c):=\left[\begin{array}{c}\delta-P_{\Delta}[\delta-\nabla_{\delta} F(\delta,c)] \nonumber
% \\
% c-P_{\mathcal{C}}[c+\nabla_{c} F(\delta,c)] \end{array}\right]
% }
% \end{equation}
where $P_{\mathcal{X}}$ denotes the projection operator on convex set $\mathcal{X}$, and $\|\mathcal{L}(\delta,c)\|$ is a commonly used measure for stationarity of the obtained solution.
In our case, $\Delta=\{\delta: x+\delta \in [0,1]^d \cap \delta \in [-\epsilon,\epsilon]^d\}$ and $\mathcal{C}=\{c: 0\le c \le \bc\}$, where $\bc$ can be an arbitrary large value.
When $\|\mathcal{L}(\delta^*,c^*)\| =0$, then the point $(\delta^*,c^*)$ is refereed as a game stationary point of the min-max problem \cite{minspm20}.
Next, we now present our main theoretical result. 
\\
{\bf Theorem 1.} \textit{Suppose Assumptions A.1 and A.2 hold and the sequence $\{\delta_t,c_t,\forall t\ge1\}$ is generated by the MinMax attack algorithm. For a given small constant $\varepsilon'$ and positive constant $\beta$, let $T(\varepsilon')$ denote the first iteration index such that the following inequality is satisfied: $T(\varepsilon'):=\min\{t|\|\mathcal{L}(\delta_t,c_t)\|^2\le\varepsilon',t\ge1\}$. Then, when the step-size and approximation error achieved by Algorithm \ref{minmax} satisfy $\alpha\sim\eta\sim\sqrt{1/T(\varepsilon')}$, there exists some constant $C$ such that $
\|\mathcal{L}(\delta_{T(\varepsilon')},c_{T(\varepsilon')})\|^2\le C/\sqrt{T(\varepsilon')}
$. }
\\
\emph{Proof}. Please see the
supplemental material (SuppMat \ref{subsec_proof}).
%see Section \ref{th.pr1}
%\PY{Why no $\beta$? $\gamma_t$ is $c_t$?}

Theorem 1 states the rate of convergence of our proposed MinMax attack algorithm when provided with sufficient stability of MINE and proper selection of the step sizes. We also remark that under the assumptions and conditions of step-sizes, this convergence rate is standard in non-convex min-max saddle point problems \cite{hibsa20}. 
\subsection{Data Augmentation using UAE}
With the proposed MinMax attack algorithm and per-sample MINE for similarity evaluation, we can generate MINE-based supervised and unsupervised adversarial examples (UAEs). Section \ref{sec_performance} will show  novel applications of
%our MinMax attack can find UAEs of training data with high attack success rate (measured by the respective training loss of perturbed and original data samples) across different datasets and unsupervised learning tasks. More importantly, by simply
MINE-based UAEs as a simple plug-in data augmentation tool to boost the model performance of several unsupervised machine learning tasks.
%augmenting the found UAEs with the training data and retraining the model (from scratch) onthe augmented dataset,
We observe significant and consistent performance improvement in data reconstruction (up to 73.5\% improvement), representation learning (up to 1.39\% increase in accuracy), and contrastive learning (1.58\% increase in accuracy). The observed performance gain can be attributed to the fact that our UAEs correspond to ``on-manifold'' data samples having low training loss but are dissimilar to the training data, causing generalization errors. Therefore, data augmentation and retraining with UAEs can improve generalization \cite{stutz2019disentangling}.   
%Consequently, in addition to evaluating the robustness of unsupervised machine learning models, UAEs can be a novel and powerful tool for data augmentation.

%Based on MinMax, we propose both unsupervised and supervised attack algorithms. For supervised attack, we demonstrate that adversarial examples crafted with large $\epsilon$ resemble nature images. Unsupervised attack can be regarded as technique of data augmentation. Furthermore, we can use UAEs to aid representation learning and contrastive learning bringing about improvements of data reconstruction and classified accuracy.
%1. Large-Lp norm attack
%2. Data augmentation, representation learning
%3. Contrastive representation

%\section{Theoretical Guarantees}
%Before showing the theoretical guarantees of the proposed PGDA algorithm, we first make the following assumptions on the properties of the problem.
%\subsection{Assumptions}

\section{Performance Evaluation}
\label{sec_performance}
In this section, we conduct extensive experiments on a variety of datasets and neural network models to demonstrate the performance of our proposed MINE-based MinMax adversarial attack algorithm and the utility of its generated UAEs for data augmentation, where a high attack success rate using UAEs suggests rich space for data augmentation to improve model performance. Codes are available at \textcolor{blue}{\url{ https://github.com/IBM/UAE}}.
%We start by describing our experiment setup and datasets in Section \ref{subsec_setup}. Section \ref{subsec_penality} compares the performance of our proposed MinMax algorithm versus the penalty-based algorithm. Section \ref{subsec_quali} shows MINE-based adversarial examples can have better visual quality than the widely used projected gradient descent (PGD) attack \cite{madry2017towards} given the same $L_\infty$ perturbation bound. We also show that data augmentation of the training data via UAEs lead to notable improvement of the test data in data reconstruction of different autoencoders (Section \ref{subsec_reconst}), representation learning of different datasets (Section \ref{subsec_representation}), and contrastive learning (Section \ref{subsec_contrastive}).

\subsection{Experiment Setup and Datasets}
\label{subsec_setup}
%Hardware; hyperprameter settings;
%We show the results of supervised and unsupervised attacks and also demonstrate applications of UAEs in this section. We run Binary Search and MinMax for supervised attack. 

\textbf{Datasets} We provide a brief summary of the datasets:
%\\
$\bullet$ \textbf{MNIST} consists of grayscale images of hand-written digits. The number of  training/test samples are 60K/10K.
%\\
$\bullet$ \textbf{SVHN} is a colar image dataset set of house numbers extracted from Google Street View images. The number of training/test samples are 73257/26302.
%\\
$\bullet$ \textbf{Fashion MNIST} contains grayscale images of 10 clothing items. The number of training/test samples are 60K/10K.
%\\
$\bullet$ \textbf{Isolet} consists of preprocessed speech data of people speaking the name of each letter of the English alphabet. %It's known as a benchmark in the feature selection literature. 
The number of training/test samples are 6238/1559.
%\\
$\bullet$ \textbf{Coil-20} contains grayscale images of 20 multi-viewed objects. The number of training/test samples are 1152/288.
%\\
$\bullet$ \textbf{Mice Protein} consists of the expression levels (features) of 77 protein modifications 
%that produced detectable signals 
in the nuclear fraction of cortex. 
%Each feature is the protein expression level. 
The number of training/test samples are 864/216.
%\\
$\bullet$ \textbf{Human Activity Recognition}
consists of sensor data collected from a smartphone for various human activities. The number of training/test samples are 4252/1492.

% \begin{table}[t]
% \vspace{-2mm}
%     \centering
%         \caption{Comparison between MinMax and penality-based algorithms in terms of supervised attack success rate (ASR) and mutual information (MI) value averaged over 1000 adversarial examples.}  
%       \begin{adjustbox}{max width=0.8\columnwidth}
%     \begin{tabular}{ccccc}
%     \toprule 
%      &\multicolumn{2}{c}{MNIST}&\multicolumn{2}{c}{CIFAR-10}\\
%     \midrule
%      & \makecell[c]{ASR} &\makecell[c] {MI}&ASR&MI\\
%      \midrule
%         Penalty-based & 100\% & 28.28 & 100\% & 13.69\\
%         MinMax        & 100\% & \textbf{51.29} & 100\% & \textbf{17.14}\\
%     \bottomrule 
%     \end{tabular}
%     \end{adjustbox}
%   \label{9000iters}    
%   \vspace{-4mm}
% \end{table}

% \begin{figure}[t]
% \vspace{-2mm}
%     \centering
%      \subfigure[MNIST]{
%     \includegraphics[scale=0.26]{images/Min_Bin_mnist.pdf}}
%     \subfigure[CIFAR-10]{
%     \includegraphics[scale=0.26]{images/Min_Bin_cifar.pdf}}
%         \vspace{-4mm}
%     \caption{
%     Mean and standard deviation of mutual information (MI) value versus attack iteration over 1000 adversarial examples. 
%     %The reported MI value of penalty-based method is %that of the current best search in $c$.
%     %Comparison of Binrary Search and MinMax in MNIST and CIFAR-10. We run 9000 iterations and record the highest MI every 1000 iterations.
%     }
%     \label{mi_vs_iters}
%       \vspace{-4mm}
% \end{figure}

\noindent \textbf{Supervised Adversarial Example Setting} Both data samples and their labels are used in the supervised setting.
We select 1000 test images classified correctly by the pretrained MNIST and CIFAR-10 deep neural network classifiers used in \cite{carlini2017towards}
and set the confidence gap parameter $\kappa=0$ for the designed attack function $f^{\text{sup}}_x$ defined in Section \ref{subsec_MINE_attack_formulation}. The attack success rate (ASR) is the fraction of the final perturbed samples leading to misclassification.

\noindent \textbf{Unsupervised Adversarial Example Setting} Only the training data samples are used in the unsupervised setting.
Their true labels are used in the post-hoc analysis for evaluating the quality of the associated unsupervised learning tasks.
All training data are used for generating UAEs individually by setting $\kappa=0$. A perturbed data sample is considered as a successful attack if its loss (relative to the original sample) is no greater than the original training loss (see Table \ref{tab:UAE_illustration}). For data augmentation, if a training sample fails to find a successful attack, we will replicate itself to maintain data balance. The ASR is measured on the training data, whereas the reported model performance is evaluated on the test data. The training performance is provided in SuppMat \ref{appen_training}.

\noindent \textbf{MinMax Algorithm Parameters}
We use consistent parameters by setting $\alpha=0.01$, $\beta = 0.1$, and $T=40$ as the default values. The vanilla MINE model \cite{belghazi2018mutual} is used in our per-sample MINE implementation. %Unless specified, we use the convolution-based approach (discussed in Section \ref{subsec_MINE_single}).
The parameter sensitivity analysis is reported in SuppMat \ref{appen_sensitivity}. 

\noindent \textbf{Computing Resource} All experiments are conducted using an Intel Xeon E5-2620v4 CPU, 125 GB RAM and a NVIDIA TITAN Xp
GPU with 12 GB RAM.

\noindent \textbf{Models and Codes} We defer the summary of the considered machine learning models to the corresponding sections. Our codes are provided in SuppMat.

%, $\beta=0.1$ for MinMax. For unsupervised attack, we set $\beta=0.1$ and attack training data for all datasets. For all of the experiments we use GD optimizer with a learning rate of 0.01.

\subsection{MinMax v.s. Penalty-based Algorithms}
\label{subsec_penality}
%Table with 9000 attack iterations over 1000 images (normal model, compare MI and ASR) - untarted. random target. 
%MI vs iteration (when f=0)
We use the same untargeted supervised attack formulation and a total of $T=9000$ iterations to compare our proposed MinMax algorithm with the penalty-based algorithm using $9$ binary search steps on MNIST and CIFAR-10. 
Table \ref{9000iters} shows that while both methods can achieve 100\% ASR, MinMax algorithm attains much higher MI values than penalty-based algorithm. The results show that the MinMax approach is more efficient in finding MINE-based adversarial examples, which can be explained by the dynamic update of the coefficient $c$ in  Algorithm \ref{minmax}.

\begin{figure}[t]
    \centering
     \subfigure[MNIST]{
    \includegraphics[scale=0.265]{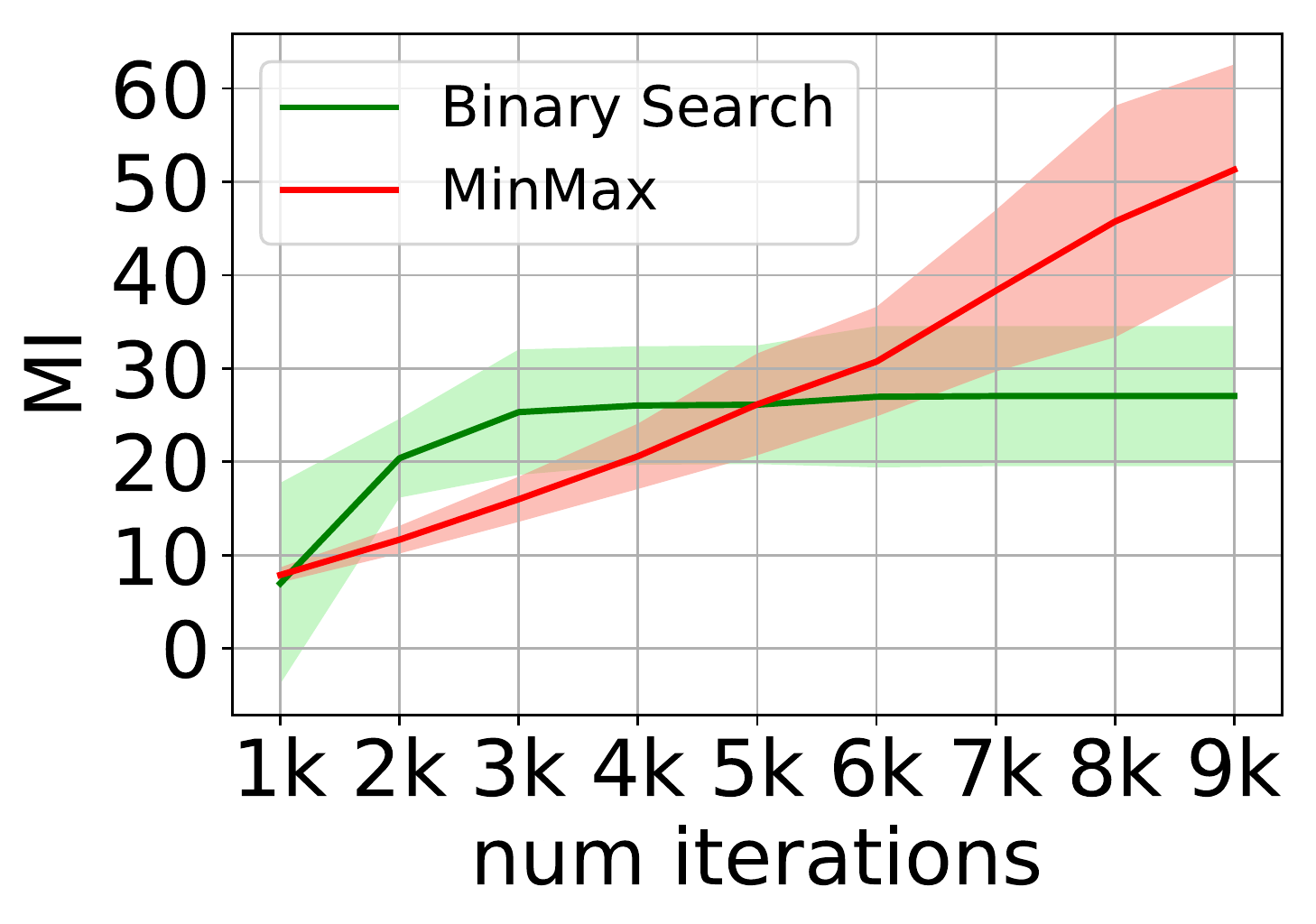}}
    \subfigure[CIFAR-10]{
    \includegraphics[scale=0.265]{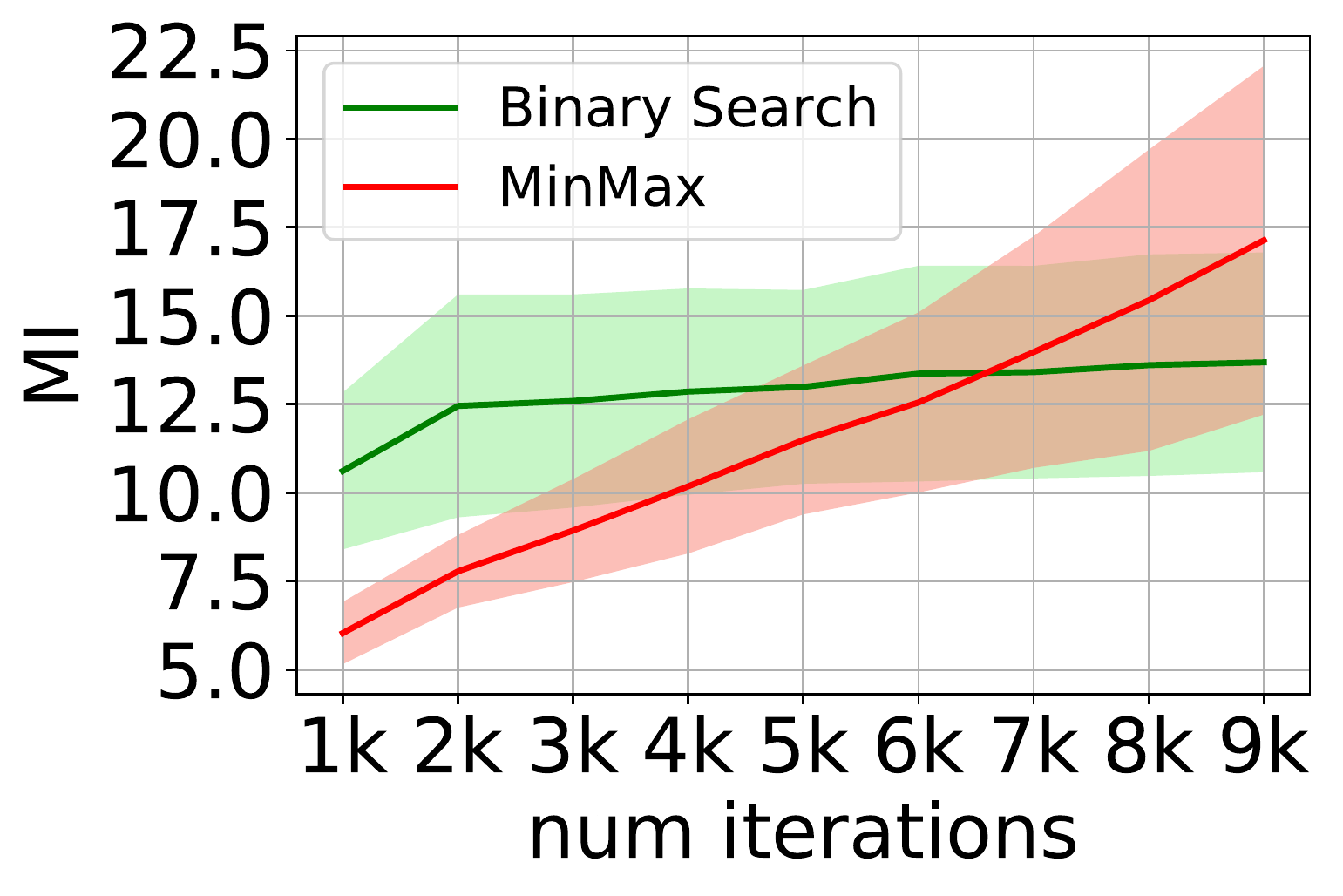}}
    \vspace{-3mm}
    \caption{
    Mean and standard deviation of mutual information (MI) value versus attack iteration over 1000 samples.}
    \label{mi_vs_iters}
    \vspace{-6mm}
\end{figure}

Figure \ref{mi_vs_iters} compares the statistics of MI values over attack iterations. One can find that as iteration count increases, MinMax algorithm can continue improving the MI value, whereas penalty-based algorithm saturates at a lower MI value due to the use of fixed coefficient $c$ in the attack process. In the remaining experiments, we will report the results using MinMax algorithm due to its efficiency.

\begin{table}[t]
    \centering
    %\vspace{-2mm}
      \begin{adjustbox}{max width=\columnwidth}
    \begin{tabular}{ccccc}
    \toprule 
     &\multicolumn{2}{c}{MNIST}&\multicolumn{2}{c}{CIFAR-10}\\
    \midrule
     & \makecell[c]{ASR} &\makecell[c] {MI}&ASR&MI\\
     \midrule
        Penalty-based & 100\% & 28.28 & 100\% & 13.69\\
        MinMax        & 100\% & \textbf{51.29} & 100\% & \textbf{17.14}\\
    \bottomrule 
    \end{tabular}
    \end{adjustbox}
    \vspace{-2mm}
     \caption{Comparison between MinMax and penality-based algorithms on MNIST and CIFAR-10 datasets in terms of attack success rate (ASR) and mutual information (MI) value averaged over 1000 adversarial examples.}
     \label{9000iters}
    \vspace{-5mm}
\end{table}

%Results indicate MinMax has high attack success rate (ASR) and high MI simultaneously shown in Table \ref{9000iters}. Besides, we show MinMax can achieve higher MI than Binary Search within 9000 iterations subjuect to attacking successful in Fig \ref{mi_vs_iters}. Noted that MI mentioned in this section is average value for 1000 test samples.%the highest MI every iterations subject to attacking successful in Fig \ref{mi_vs_iters}.
\subsection{Qualitative Visual Comparison}
\label{subsec_quali}
\begin{figure}[t]
\vspace{0mm}
    \centering
    \subfigure[]{\includegraphics[scale=0.128]{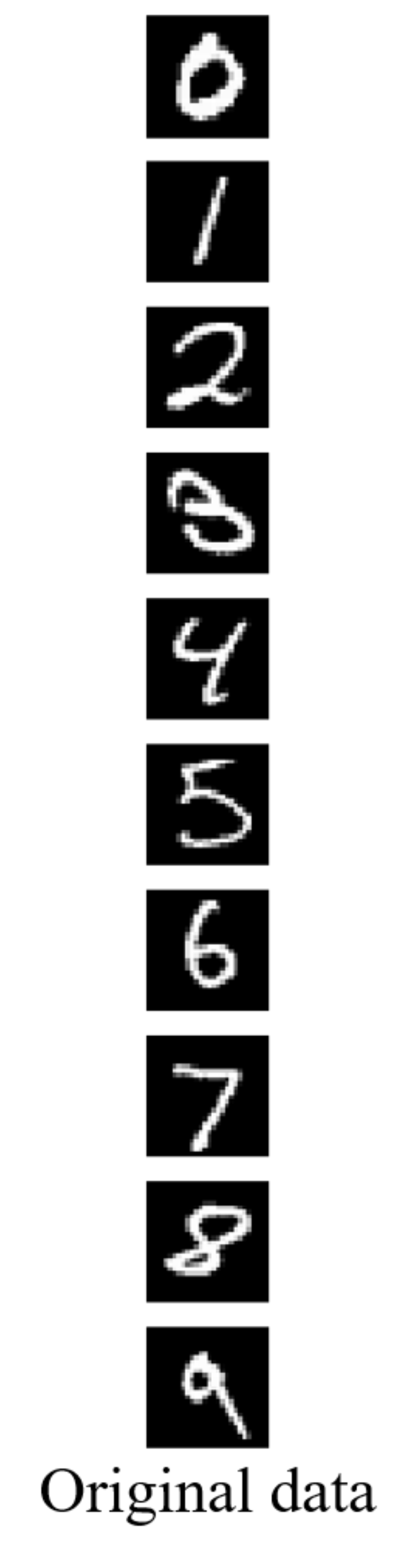}}
     \subfigure[PGD attack]{
    \includegraphics[scale=0.16]{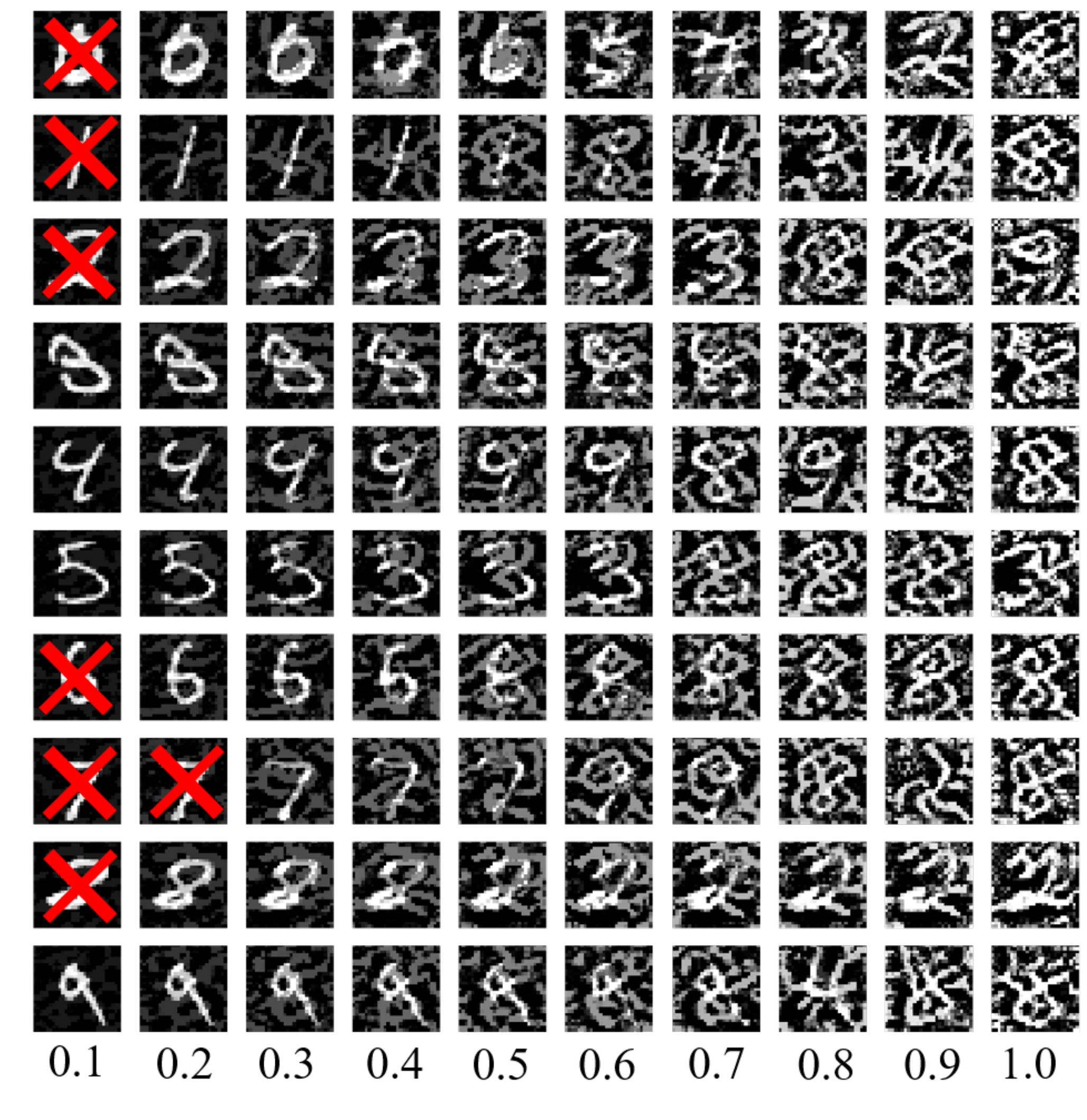}}
    \subfigure[MinMax attack]{
    \includegraphics[scale=0.16]{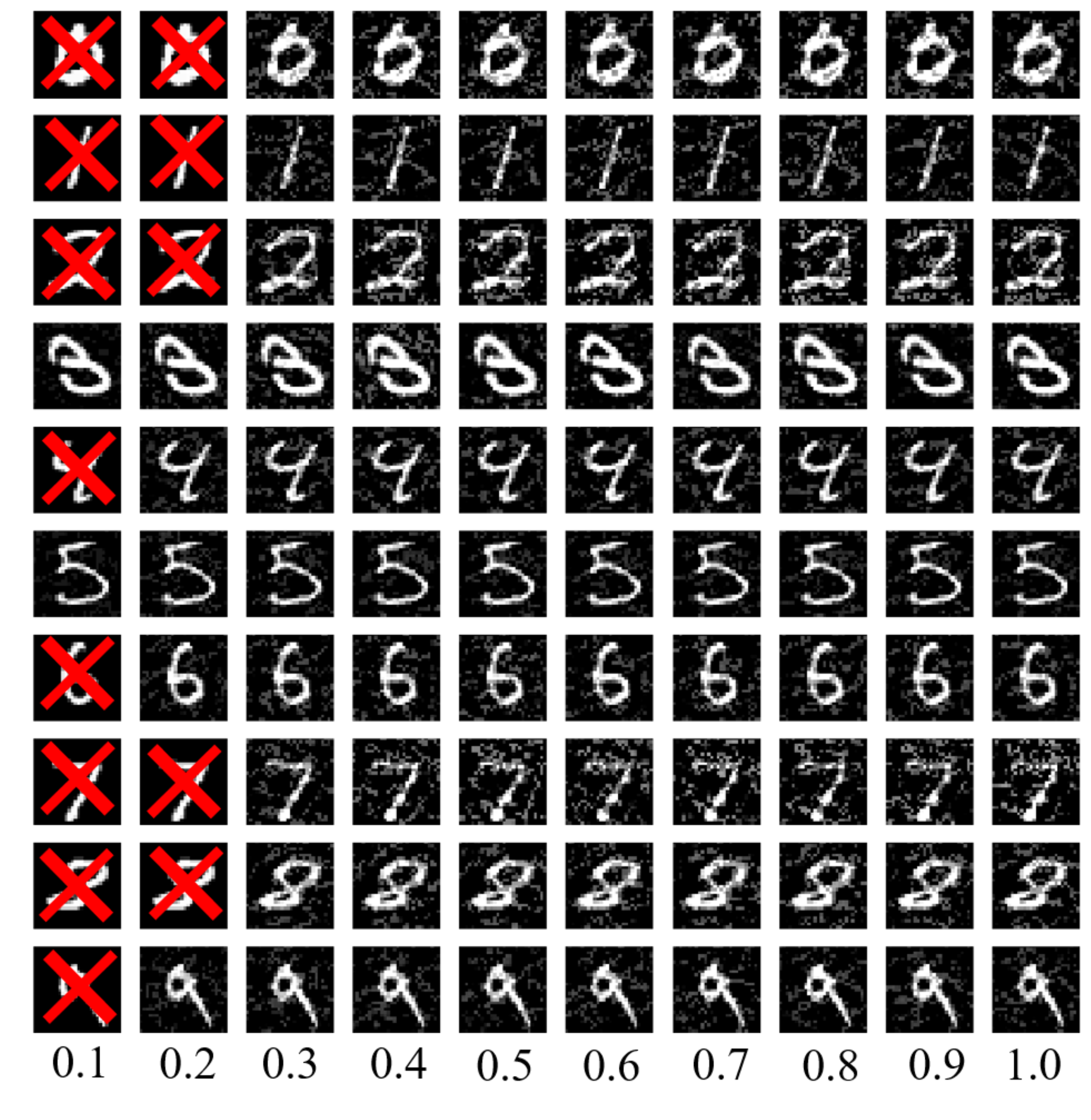}}
        \vspace{-4mm}
    \caption{Comparison of untargeted supervised adversarial examples on MNIST. The unsuccessful adversarial examples are marked with red crosses.  Each column corresponds to different $\epsilon$ values ($L_\infty$-norm perturbation bound) ranging from $0.1$ to $1.0$.
    Each row shows the adversarial examples of an original sample. MinMax attack using MINE yields adversarial examples with better visual quality than PGD attack, especially for large $\epsilon$ values. }
    \label{Fig_visual_MNIST}
    \vspace{-5mm}
\end{figure}

Figure \ref{Fig_visual_MNIST} presents a visual comparison of MNIST supervised adversarial examples 
crafted by MinMax attack and the PGD attack with 100 iterations \cite{madry2017towards}
given different $\epsilon$ values governing the $L_\infty$ perturbation bound. The main difference is that MinMax attack uses MINE as an additional similarity regulation while PGD attack only uses $L_\infty$ norm. Given the same $\epsilon$ value, MinMax attack yields adversarial examples with better visual quality. 
The results validate the importance of MINE as an effective similarity metric. In contrast, PGD attack aims to make full use of the $L_\infty$ perturbation bound and attempts to modify every data dimension, giving rise to lower-quality adversarial examples. Similar results are observed for adversarially robust models \cite{madry2017towards,zhang2019theoretically}, as shown in SuppMat \ref{appen_visual_robust}.

%\begin{wrapfigure}{R}{0.57\textwidth}
%\end{wrapfigure}

Moreover,  the results also suggest that for MINE-based attacks, the $L_\infty$ norm constraint on the perturbation is not critical for the resulting visual quality, which can be explained by the fact that 
MI is a fundamental information-theoretic similarity measure.
When performing MINE-based attacks, we suggest not using the $L_\infty$ norm constraint
(by setting $\epsilon=1$) so that the algorithm can fully leverage the power of MI to find a more diverse set of adversarial examples.

%Adversarial examples are likely normal images in human eyes under attack with large $\epsilon$ because perturbations are still restricted by maximization of mutual information.
%Large eps, ours is better than others

Next, we study three different unsupervised learning tasks. We only use the training samples and the associated training loss to generate UAEs. The post-hoc analysis reports the performance on the test data and the downstream classification accuracy. We report their improved adversarial robustness after data augmentation with MINE-UAEs in SuppMat \ref{subsec_robustness}.

\subsection{UAE Improves Data Reconstruction}
\label{subsec_reconst}

Data reconstruction using an autoencoder $\Phi(\cdot)$ that learns to encode and decode the raw data through latent representations is a standard unsupervised learning task. Here we use the default implementation of the following four autoencoders to generate UAEs based on the training data samples of MNIST and SVHN for data augmentation, retrain the model from scratch on the augmented dataset, and  report the resulting reconstruction error on the original test set. The results of larger-scale datasets (CIFAR-10 and Tiny-ImageNet) are reported in SuppMat \ref{appen_recon_large}.
All autoencoders use the $L_2$ reconstruction loss defined as $\|x-\Phi(x)\|_2$. We provide more details about the model retraining in SuppMat \ref{appen_training}.

%The four autoencoders are summarized below.\\
\noindent $\bullet$ \textbf{Dense Autoencoder \cite{cavallari2018unsupervised}:} The encoder and decoder have 1 dense layer separately and the latent dimension is 128/256 for MNIST/SVHN.\\
$\bullet$ \textbf{Sparse Autoencoder:} It has a sparsity enforcer ($L_1$ penalty on the training loss) that directs a network with a single hidden layer to learn the latent representations minimizing the error in reproducing the input while limiting the number of code words for reconstruction. We use the same architecture as Dense Autoencoder for MNIST and SVHN. 
%\PY{For MNIST, we use the default model. For SVHN, we apply the same MNIST model and change the latent dimension to be XXX.}
%\PY{The latent dimension varies by datasets.}
\\
$\bullet$ \textbf{Convolutional Autoencoder\footnote{We use \url{https://github.com/shibuiwilliam/Keras_Autoencoder}}:}  The encoder uses convolution+relu+pooling layers.
The decoder has reversed layer order with
the pooling layer replaced by an upsampling layer. %Convolution layers can use input data like images without any modification. 
\\
$\bullet$ \textbf{Adversarial Autoencoder \cite{makhzani2015adversarial}:} It is composed of an encoder, a decoder and a discriminator.
The rationale is to force the distribution of the encoded values to be similar to the prior data distribution. 

%As a result, the decoder learns only the mapping from the prior distribution to data distribution.

\begin{table*}[t]
%\vspace{-2mm}
  \centering
  %\vspace{-2mm}
  \begin{adjustbox}{max width=0.8\textwidth}
  \begin{tabular}{c|ccccc|cccc}
    \toprule
     \multicolumn{10}{c}{MNIST} \\
     \midrule
      & \multicolumn{5}{c}{Reconstruction Error (test set)}&  \multicolumn{4}{|c}{ASR (training set)}\\
     \midrule
         Autoencoder & Original & \makecell[c]{MINE-UAE} & $L_{2}$-UAE & \makecell[c]{GA\\ ($\sigma=0.01$)} &\makecell[c]{GA\\ ($\sigma=10^{-3}$)}& \makecell[c]{MINE-UAE } &   $L_{2}$-UAE & \makecell[c]{GA\\ ($\sigma=0.01$)}&\makecell[c]{GA\\ ($\sigma=10^{-3}$)}   \\
    \midrule
     Sparse        & 0.00561 & \makecell[c]{\textbf{0.00243}\\ (\textcolor{ForestGreen}{$\uparrow 56.7\%$})} & \makecell[c]{0.00348 \\ (\textcolor{ForestGreen}{$\uparrow 38.0\%$})}  & \makecell[c]{0.00280$\pm$2.60e-05\\(\textcolor{ForestGreen}{$\uparrow50.1\%$})} & \makecell[c]{0.00280$\pm$3.71e-05\\(\textcolor{ForestGreen}{$\uparrow50.1\%$})} & 100\%   & 99.18\% &54.10\% & 63.95\%\\ 
     Dense          & 0.00258 & \makecell[c]{\textbf{0.00228}\\ (\textcolor{ForestGreen}{$\uparrow 11.6\%$})} & \makecell[c]{0.00286\\ (\textcolor{red}{$\downarrow 6.0\%$})}  & \makecell[c]{0.00244$\pm$0.00014\\($\textcolor{ForestGreen}{\uparrow5.4\%}$)} & \makecell[c]{0.00238$\pm$0.00012\\(\textcolor{ForestGreen}{$\uparrow7.8\%$})}  & 92.99\% & 99.94\% &48.53\%&58.47\%\\
     \makecell[c]{Convolutional} & 0.00294  & \makecell[c]{\textbf{0.00256}\\ (\textcolor{ForestGreen}{$\uparrow 12.9\%$})} & \makecell[c]{0.00364\\ (\textcolor{red}{$\downarrow 23.8\%$})} & \makecell[c]{0.00301$\pm$0.00011\\ (\textcolor{red}{$\downarrow2.4\%$})}& \makecell[c]{0.00304$\pm$0.00015\\(\textcolor{red}{$\downarrow3.4\%$})}& 99.86\% & 99.61\%&68.71\%&99.61\% \\
     \makecell[c]{Adversarial}   & 0.04785 & \makecell[c]{\textbf{0.04581}\\ (\textcolor{ForestGreen}{$\uparrow 4.3\%$})} & \makecell[c]{0.06098\\ (\textcolor{red}{$\downarrow 27.4\%$})}  & \makecell[c]{0.05793$\pm$0.00501\\(\textcolor{red}{$\downarrow21\%$})} &\makecell[c]{0.05544$\pm$0.00567\\(\textcolor{red}{$\downarrow15.86\%$})} &98.46\% &43.54\%&99.79\%&99.83\%\\
     %\midrule
     \toprule
     \multicolumn{10}{c}{SVHN} \\
     \midrule
     Sparse         & 0.00887 & \makecell[c]{\textbf{0.00235}\\ (\textcolor{ForestGreen}{$\uparrow 73.5\%$})} & \makecell[c]{0.00315\\ (\textcolor{ForestGreen}{$\uparrow 64.5\%$})}   &  \makecell[c]{0.00301$\pm$0.00137\\(\textcolor{ForestGreen}{$\uparrow66.1\%$})}  & \makecell[c]{0.00293$\pm$0.00078\\(\textcolor{ForestGreen}{$\uparrow67.4\%$})}& 100\%   & 72.16\% & 72.42\% & 79.92\% \\
     Dense         & 0.00659 & \makecell[c]{\textbf{0.00421}\\ (\textcolor{ForestGreen}{$\uparrow 36.1\%$})} & \makecell[c]{0.00550\\ (\textcolor{ForestGreen}{$\uparrow 16.5\%$})}   &        \makecell[c]{0.00858$\pm$0.00232\\(\textcolor{red}{$\downarrow30.2\%$})}
      & \makecell[c]{0.00860$\pm$0.00190\\(\textcolor{red}{$\downarrow30.5\%$})} & 99.99\% & 82.65\% & 92.3\% & 93.92\% \\
     \makecell[c]{Convolutional} & 0.00128 & \makecell[c]{\textbf{0.00095}\\ (\textcolor{ForestGreen}{$\uparrow 25.8\%$})} & \makecell[c]{0.00121\\ (\textcolor{ForestGreen}{$\uparrow 5.5\%$})}  & \makecell[c]{0.00098 $\pm$ 3.77e-05\\ (\textcolor{ForestGreen}{$\uparrow 25.4\%$})} & \makecell[c]{0.00104$\pm$7.41e-05\\(\textcolor{ForestGreen}{$\uparrow18.8\%$})} & 100\%   & 56\%  &96.40\% &99.24\%\\
     \makecell[c]{Adversarial}   & 0.00173 & \makecell[c]{\textbf{0.00129}\\ (\textcolor{ForestGreen}{$\uparrow 25.4\%$})} & \makecell[c]{0.00181\\ (\textcolor{red}{$\downarrow 27.4\%$})} & \makecell[c]{0.00161$\pm$0.00061\\(\textcolor{ForestGreen}{$\uparrow6.9\%$})}& \makecell[c]{0.00130$\pm$0.00037\\(\textcolor{ForestGreen}{$\uparrow24.9\%$})} & 94.82\% & 58.98\%&97.31\%&99.85\% \\
     %\midrule
    \bottomrule
\end{tabular}
\end{adjustbox}
\vspace{-2mm}
\caption{Comparison of data reconstruction by retraining the autoencoder on UAE-augmented data. The error is the average $L_2$ reconstruction loss of the test set. The improvement (in \textcolor{ForestGreen}{green}/\textcolor{red}{red}) is relative to the original model. The 
  attack success rate (ASR) is the fraction of augmented training data having smaller reconstruction loss than the original loss (see Table \ref{tab:UAE_illustration} for definition).  }
\label{recons error}
   \vspace*{-6mm}
\end{table*}

%Autoencoder

We also compare the performance of our proposed MINE-based UAE (MINE-UAE) with two baselines: (i) \textit{$L_2$-UAE} that replaces the objective of minimizing $I_\Theta(x,x+\delta)$ with maximizing the $L_2$ reconstruction loss $\|x-\Phi(x+\delta)\|_2$ in the MinMax attack algorithm while keeping the same attack success criterion; (ii) \textit{Gaussian augmentation} (GA) that adds zero-mean Gaussian noise with a diagonal covariance matrix of the same constant
$\sigma^2$ to the training data.

Table \ref{recons error} shows the reconstruction loss and the ASR. The improvement of reconstruction error is measured with respect to the reconstruction loss of the original model (i.e., without data augmentation). We find that MINE-UAE can attain much higher ASR than $L_2$-UAE and GA in most cases. More importantly, data augmentation using MINE-UAE achieves consistent and significant reconstruction performance improvement across all models and datasets (up to 
$56.7\%$ on MNIST and up to $73.5\%$ on SVHN), validating the  effectiveness of MINE-UAE for data augmentation. On the other hand, in several cases $L_2$-UAE and GA lead to notable performance degradation. The results suggest that MINE-UAE can be an effective plug-in data augmentation tool for boosting the performance of unsupervised machine learning models.
%, 
%as it simply uses the training data and the original model to generate UAEs for model retraining.
Table \ref{several augs} demonstrates UAEs can  further improve data reconstruction when the original model already involves conventional augmented training data such as flip, rotation, and Gaussian noise. The augmentation setup is given in SuppMat \ref{appen_add_aug}. We also show the run time analysis of different augmentation method in SuppMat \ref{append_runtime}.

%\vspace{-2mm}
\begin{table}[t]
\centering
    %\vspace*{-1mm}
    \begin{adjustbox}{max width=.99\columnwidth}
    \begin{tabular}{ccc}
    \toprule 
     \multicolumn{3}{c}{SVNH - Convolutional AE}\\
    \midrule
     Augmentation & \makecell[c]{Aug. (test set)} & Aug.+MINE-UAE (test set) \\
     \midrule
        Flip + Rotation & 0.00285 & \makecell[c]{\textbf{0.00107} \textcolor{ForestGreen}{ ($\uparrow 62.46\%$)}} \\      \midrule
        \makecell[c]{Gaussian noise ($\sigma=0.01$)} &  0.00107&\makecell[c]{\textbf{0.00095} \textcolor{ForestGreen}{ ($\uparrow 11.21\%$)}}\\
         \midrule
        \makecell[c]{Flip + Rotation + Gaussian noise}& 0.00307 & \makecell[c]{\textbf{0.00099} \textcolor{ForestGreen}{ ($\uparrow 67.75\%$)}}\\
    \bottomrule 
    \end{tabular}
    \end{adjustbox}
    \vspace{-2mm}
    \caption{Performance of data reconstruction when retraining with MINE-UAE and additional augmented training data.}
     \label{several augs}
    \vspace*{-3mm}
\end{table}

\begin{table}
    \centering
    %\vspace{-2mm}
    \begin{adjustbox}{max width=1\columnwidth}
    \begin{tabular}{c|cc|cc|c}
    \toprule 
     &\multicolumn{2}{|c|}{Reconstruction Error (test set)}&  \multicolumn{2}{|c|}{Accuracy (test set)}& ASR\\
    \midrule
     Dataset & Original & MINE-UAE & Original & MINE-UAE & MINE-UAE\\
    \midrule
        MNIST         & 0.01170 & \textbf{0.01142} (\textcolor{ForestGreen}{$\uparrow 2.4\%$}) & 94.97\% & 95.41\% & 99.98\% \\
        Fashion MMIST & 0.01307 & \textbf{0.01254} (\textcolor{ForestGreen}{$\uparrow 4.1\%$}) & 84.92\% & 85.24\% & 99.99\% \\
        Isolet        & 0.01200 & \textbf{0.01159} (\textcolor{ForestGreen}{$\uparrow 3.4\%$}) & 81.98\% & 82.93\% &   100\% \\
        Coil-20       & \textbf{0.00693} & 0.01374 (\textcolor{red}{$\downarrow 98.3\%$}) & 98.96\% & 96.88\% &  9.21\% \\
        Mice Protein  & 0.00651 & \textbf{0.00611} (\textcolor{ForestGreen}{$\uparrow 6.1\%$}) & 89.81\% & 91.2\%  & 40.24\% \\
        Activity      & 0.00337 & \textbf{0.00300} (\textcolor{ForestGreen}{$\uparrow 11.0\%$}) & 83.38\% & 84.45\% & 96.52\% \\
    \bottomrule 
    \end{tabular}
    \end{adjustbox}
    \vspace{-2mm}
    \caption{Performance of representation learning by the concrete autoencoder and the resulting classification accuracy. The degradation on Coil-20 is explained in Section \ref{subsec_representation}. }  
    \label{concrete}
    \vspace*{-5mm}
\end{table}

\subsection{UAE Improves Representation Learning}
\label{subsec_representation}

The concrete autoencoder \cite{balin2019concrete} is 
an unsupervised feature selection method which recognizes a subset of the most informative features
through an additional \emph{concrete select layer} with $M$ nodes in the encoder for data reconstruction.
We apply MINE-UAE for data augmentation and use the same post-hoc classification evaluation procedure as in 
\cite{balin2019concrete}.
%for the learned representations.
%which 
%passes the selected features to an extremely randomized tree classification model \cite{geurts2006extremely}.

The six datasets and the resulting classification accuracy are reported in Table \ref{concrete}. We select $M=50$ features for every dataset except for Mice Protein (we set $M=10$) owing to its small data dimension.
MINE-UAE can attain up to 11\% improvement for data reconstruction and up to 1.39\% increase in accuracy among 5 out of 6 datasets, corroborating the utility of MINE-UAE in representation learning and feature selection.
The exception is Coil-20. A closer inspection shows that MINE-UAE has low ASR ($<$10\%) for Coil-20 and the training loss after data augmentation is significantly higher than the original training loss (see SuppMat \ref{appen_training}). Therefore, we conclude that the degraded performance in Coil-20 after data augmentation is likely due to the limitation of feature selection protocol and the model learning capacity.

% \begin{table}[t]
% \vspace{-2mm}
%     \centering
%     \caption{Performance evaluation of representation learning by the concrete autoencoder and the resulting classification accuracy. The observed degradation on Coil-20 is explained in Section \ref{subsec_representation}. }     
%     \label{concrete}
%     \begin{adjustbox}{max width=\columnwidth}
%     \begin{tabular}{c|cc|cc|c}
%     \toprule 
%      &\multicolumn{2}{|c|}{Reconstruction Error (test set)}&  \multicolumn{2}{|c|}{Accuracy (test set)}& ASR\\
%     \midrule
%      Dataset & Original & MINE-UAE & Original & MINE-UAE & MINE-UAE\\
%     \midrule
%         MNIST         & 0.01170 & \textbf{0.01142} (\textcolor{ForestGreen}{$\uparrow 2.4\%$}) & 94.97\% & 95.41\% & 99.98\% \\
%         Fashion MMIST & 0.01307 & \textbf{0.01254} (\textcolor{ForestGreen}{$\uparrow 4.1\%$}) & 84.92\% & 85.24\% & 99.99\% \\
%         Isolet        & 0.01200 & \textbf{0.01159} (\textcolor{ForestGreen}{$\uparrow 3.4\%$}) & 81.98\% & 82.93\% &   100\% \\
%         Coil-20       & \textbf{0.00693} & 0.01374 (\textcolor{red}{$\downarrow 98.3\%$}) & 98.96\% & 96.88\% &  9.21\% \\
%         Mice Protein  & 0.00651 & \textbf{0.00611} (\textcolor{ForestGreen}{$\uparrow 6.1\%$}) & 89.81\% & 91.2\%  & 40.24\% \\
%         Activity      & 0.00337 & \textbf{0.00300} (\textcolor{ForestGreen}{$\uparrow 11.0\%$}) & 83.38\% & 84.45\% & 96.52\% \\
%     \bottomrule 
%     \end{tabular}
%     \end{adjustbox}
%     \vspace{-5mm}
% \end{table}

\subsection{UAE Improves Contrastive Learning}
\label{subsec_contrastive}

\begin{table}[t]
%\begin{wraptable}{r}{6.1cm}
    \centering
    %\vspace{-1mm}
    \begin{adjustbox}{max width=0.99\columnwidth}
    \begin{tabular}{cccc}
    \toprule 
     \multicolumn{4}{c}{CIFAR-10}\\
    \midrule
     Model & \makecell[c]{Loss (test set)} & Accuracy (test set) & ASR\\
     \midrule
        Original & 0.29010 & 91.30\% & - \\
        MINE-UAE & \makecell[c]{\textbf{0.26755} \textcolor{ForestGreen}{ ($\uparrow 7.8\%$)}} & \textbf{+1.58\%} & 100\% \\
    CLAE \cite{ho2020contrastive} & - & +0.05\% & - \\
    \bottomrule 
    \end{tabular}
    \end{adjustbox}
    \vspace{-3mm}
    \caption{Comparison of contrastive loss and the resulting accuracy on CIFAR-10 using SimCLR \cite{chen2020simple} (ResNet-18 with batch size = 512). The attack success rate (ASR) is the fraction of augmented training data having smaller contrastive loss than original loss.  For CLAE \cite{ho2020contrastive}, we use the reported accuracy improvement (it shows negative gain in our implementation), though its base SimCLR model only has 83.27\% test accuracy.}    
    \label{constrastive}
    \vspace*{-5mm}
\end{table}
%\end{wraptable}

The SimCLR algorithm  \cite{chen2020simple} is a popular contrastive learning framework for visual representations. It uses self-supervised data modifications  to efficiently improve several downstream image classification tasks. %It consists of an encoder used to learn representations and non-linear projection head used to map latent space to a embedding vector. To learn representations, the encoder maximizes the similarity between applying 2 augmentation functions on the same image and minimize on the different image. It learns high quality of the representation and improves several image classification tasks. 
We use the default implementation of SimCLR on CIFAR-10 and generate MINE-UAEs using the training data and the defined training loss for SimCLR. Table \ref{constrastive} shows the loss, ASR and the resulting classification accuracy by taining a linear head on the learned representations. We find that using MINE-UAE for additional data augmentation and model retraining can yield 7.8\% improvement in contrastive loss and 1.58\% increase in classification accuracy.
Comparing to \cite{ho2020contrastive} using adversarial examples to improve SimCLR (named CLAE), the accuracy increase of MINE-UAE is 30x higher.
Moreover, MINE-UAE data augmentation also significantly improves adversarial robustness (see SuppMat \ref{subsec_robustness}).

\section{Conclusion}
In this paper, we propose a novel framework for studying adversarial examples in unsupervised learning tasks, based on our developed  per-sample mutual information neural estimator as an information-theoretic similarity measure. We also propose a new MinMax algorithm for efficient generation of MINE-based supervised and unsupervised adversarial examples and establish its convergence guarantees. As a novel application, we show that MINE-based UAEs can be used as a simple yet effective plug-in data augmentation tool and achieve significant  performance gains in data reconstruction, representation learning, and contrastive learning. 

\clearpage
\newpage

\section*{Acknowledgments}
This work was primarily done during Chia-Yi's  visit at IBM Research.
Chia-Yi Hsu and Chia-Mu Yu were supported by MOST 110-2636-E-009-018, and we also thank National Center for High-performance Computing (NCHC) of National Applied Research Laboratories (NARLabs) in Taiwan for providing computational and storage resources.

\bibliography{reference.bbl, adversarial_learning.bbl}

\clearpage
\newpage
\onecolumn
\section{Supplementary Material}
\subsection{Codes}
Our codes are provided as a zip file for review.

\subsection{More Details on Per-sample MINE}
\label{appen_MINE}

\subsubsection{Random Sampling}
We reshape an input data sample as a vector $x \in \mathbb{R}^{d}$ and independently generate $K$ Gaussian random matrices $\{M_{k}\}_{k=1}^K$, where $M_{k} \in \mathbb{R}^{d' \times d}$. Each entry in $M_k$ is an i.i.d zero-mean Gaussian random variable with standard deviation $1/d'$. The compressed samples $\{x_k\}_{k=1}^K$ of $x$ is defined as $x_k = M_k x$. Similarly, the same random sampling procedure is used on $x+\delta$ to obtain its compressed samples  $\{(x+\delta)_k\}_{k=1}^K$. In our implementation, we set $d'=128$ and $K=500$. 
%We do $x \times \{m_{i}\}_{i=1}^{i=n} = s_{i},\; s_{i} \in R^{1\times k}$ and use $\{s_{i}\}_{i=1}^{i=n}$ as batch data.
%@Chia-Yi put the sampling formulation here in math

\subsubsection{Convolution Layer Output}
Given a data sample $x$, we fetch its output of the $1^{st}$ convolutional layer, denoted by $conv(x)$. The data dimension is $d' \times K$, where $K$ is the number of filters (feature maps) and $d'$ is the (flattend) dimension of the feature map. Each filter is regarded as a compressed sample denoted by $conv(x)_k$. %and we reshape it as [number of filters, $n$, $n$, 1]. 
Algorithm \ref{MINE} summarizes the proposed approach, where the function $T_\theta$ is parameterized by a neural network $\theta$ based on the Donsker-Varadhan representation theorem \cite{donsker1983asymptotic}, and $T_I$ is the number of iterations for training the MI neural estimator $I(\theta)$.

\begin{algorithm}[H]
\caption{Per-sample MINE via Convolution}\label{MINE}
\begin{algorithmic}[1]
\State  {\bfseries Require:} input sample $x$, perturbed sample $x+\delta$, 1st convolution layer output $conv(\cdot)$, MI neural estimator $I(\theta)$
%\Ensure $\nu(\theta)$: MI
\State Initialize neural network parameters $\theta$
\State Get $\{conv(x)_{k}\}_{k=1}^K$ and $\{conv(x+\delta)_{k}\}_{k=1}^K$ via $1^{st}$ convolution layer
%\State $X_{n} \leftarrow$  Reshape $conv(x)_{k}$
%\State $Z_{n} \leftarrow$ Reshape $conv(x+\delta)_{k}$  
\For {$t$ in $T_I$ iterations}
    \State Take $K$ samples from the joint distribution:
    $\{ conv(x)_{k}$, $ conv(x+\delta)_{k} \}_{k=1}^K$
    \State Shuffle $K$ samples from $conv(x+\delta)$ marginal distribution:
    $\{conv(x+\delta)_{(k)}\}_{k=1}^K$
    \State Evaluate the mutual information estimate
    $I(\theta) \leftarrow \frac{1}{K}\sum_{k=1}^K T_{\theta} (conv(x)_{k}, conv(x+\delta)_{k})-
    \log \left( \frac{1}{K}  \sum_{k=1}^K \exp [  T_{\theta}(conv(x)_{k}, conv(x+\delta)_{(k)} ) ]  \right)$
    \State $\theta \leftarrow \theta + \nabla_{\theta} I(\theta)$
\EndFor
\State  {\bfseries Return} $I(\theta)$
\end{algorithmic}
\end{algorithm}

\subsection{Ablation Study of $K$ for Random Sampling of Per-sample MINE}\label{append_K}
We follow the same setting as in Table \ref{fid_vs_kid} on CIFAR-10 and report the average persample-MINE, FID and KID results over 1000 samples when varying the value $K$ in random sampling. The results in Table \ref{tab:K_VS_fid_mi_kid} show that both KID and FID scores decrease (meaning the generated adversarial examples are closer to the training data’s representations) as $K$ increases, and they saturate when $K$ is greater than 500. Similarly, the MI values become stable when $K$ is greater than 500. 
\begin{table*}[h]
    \centering
    \begin{tabular}{c|ccccccccc}
    \toprule
    \multicolumn{10}{c}{CIFAR-10}\\
    \toprule
         K& 50 & 100 & 200 & 300 & 400 & 500 & 600 & 700 & 800 \\
     \midrule
         MI& 1.35 &  1.53 &  1.86 & 2.05 & 2.13 &  2.32 & 2.33 & 2.35 & 2.38 \\
         FID&226.63 & 213.01 & 207.81 & 205.94 & 203.73 & 200.12 & 200.02 & 198.75 & 196.57\\
         KID&14.7  & 12.20  &  10.8  & 10.02  &  9.59  &  8.78  &  8.81  &  8.22  &  8.29\\
    \toprule
    \end{tabular}
    \caption{Comparison of MI, FID and KID when varying the value $K$ in random sampling.}
    \label{tab:K_VS_fid_mi_kid}
\end{table*}

\subsection{Additional Visual Comparisons}

\subsubsection{Visual Comparison of Supervised Adversarial Examples with $\epsilon = 0.03$}
\begin{figure}[t]
    \centering
    \includegraphics[scale=0.45]{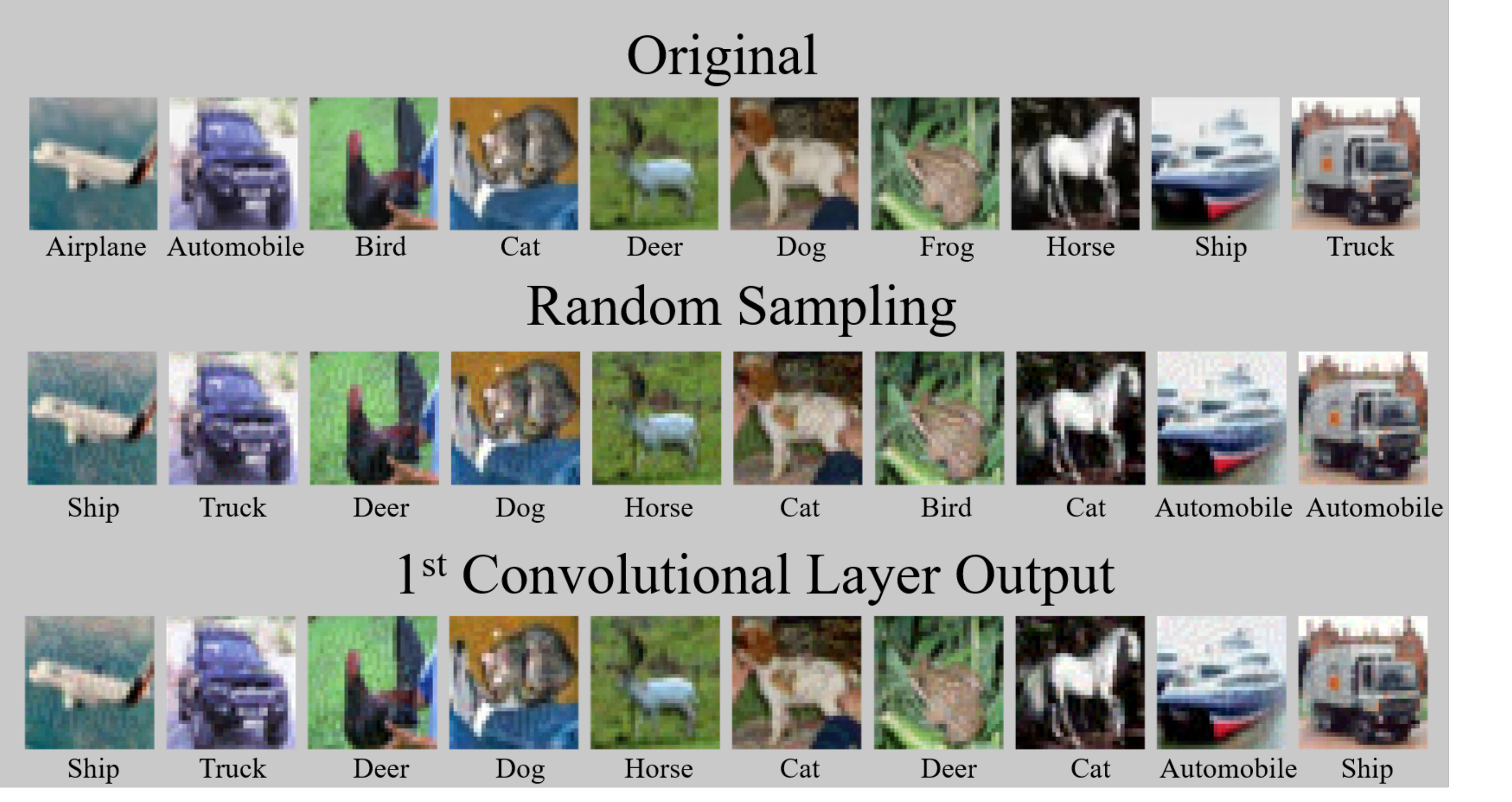}
    \caption{Visual comparison of MINE-based supervised adversarial examples (untargeted attack with $\epsilon=0.03$) on CIFAR-10. Both random sampling and convolution output can be used to craft adversarial examples with high similarity. The label below each data sample indicates the predicted label, where the predicted label of the original image is the ground-truth label.
    %We generate batches of images for MINE with different methods. 
     }
     \label{random_conv_0.03}
\end{figure}

Similar to the setting in Figure \ref{random_conv},  we compare the MINE-based supervised adversarial examples on CIFAR-10 with the $L_\infty$ constraint $\epsilon = 0.03$ (instead of  $\epsilon = 1$) in Figure \ref{random_conv_0.03}.

\subsubsection{Visual Comparison of Unsupervised Adversarial Examples}
\begin{figure}[t]
    \centering
    \includegraphics[scale=0.45]{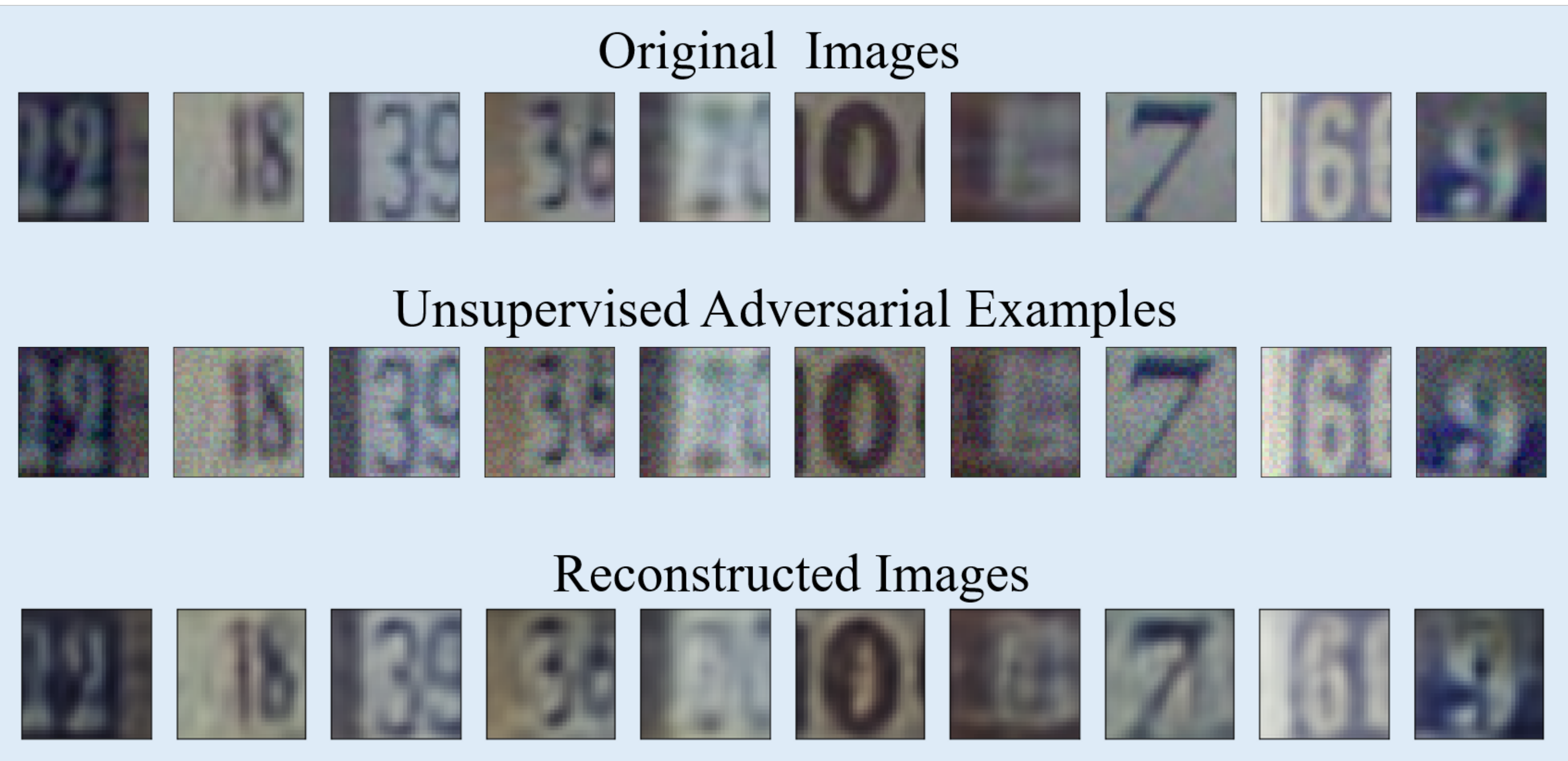}
    \caption{Visual comparison of MINE-based unsupervised adversarial examples on SVHN using convolutional autoencoder. }
     \label{UAEs_svhn}
\end{figure}

Figure \ref{UAEs_svhn} shows the generated MINE-UAEs with $\epsilon =1$ on SVHN using the convolutional autoencoder. 
We pick the 10 images such that their reconstruction loss is no greater than that of the original image, while they have the top-10 perturbation level measured by the $L_2$ norm on the perturbation $\|\delta^*\|_2$.

\clearpage

\subsection{Binary Search for Penalty-based Attack Algorithm}
\label{appen_binary}

Algorithm \ref{binarysearch} summarizes the binary search strategy on the regularization coefficient $c$ in the penalty-based methods. The search procedure follows the implementation in \cite{carlini2017towards}, which updates $c$ per search step using a pair of pre-defined upper and lower bounds.
The reported MI value of penalty-based method in Figure \ref{mi_vs_iters} is that of the current best search in $c$, where each search step takes 1000 iterations (i.e., $B=9$ and $T'=1000$). 

%\PY{update notation, $f_x^+$}

\begin{algorithm}[H]
\caption{Binary Search}\label{binarysearch}
\begin{algorithmic}[1]
\State  {\bfseries Require:} data sample $x$, attack criterion $f_{x}(\cdot)$, mutual information $I_{\Theta}(x,x+\delta)$ and step sizes $\alpha$, perturbation bound $\epsilon$, number of iterations $T'$ in each search, total number of binary search steps $B$

%\State for all capsule $i$ in layer $l$ and capsule $j$ in layer $(l+1)$: $b_{ij} \gets 0$.
\State Initialization: $lb=10^{-3}$ (lower bound on $c$), $ub=10^9$ (upper bound on $c$), $c=10^{-3}$, $I^*_{\Theta}=-\infty$, $\delta^*=\text{null}$
\For{$b$ in $B$ binary search steps}
\For{$t$ in $T'$ iterations}
\State $\delta_{t+1} = \delta_{t} - \alpha \cdot (c \cdot \nabla f^{+}_{x}(x+\delta_{t}) - \nabla I_{\Theta}(x,x+\delta_{t})) $
\State Project $\delta_{t+1}$ to $[\epsilon, -\epsilon]$ via clipping
\State Project $x+\delta_{t+1}$ to $[0, 1]$ via clipping
\State Compute $I_{\Theta}(x, x + \delta_{t+1})$
\If {$f_x(x+\delta_{t+1}) \leq 0$ and $I_{\Theta}(x, x+\delta_{t+1})>$ $I^*_{\Theta}$}
\State $\delta^* = \delta_{t+1}$
\State $I^*_\Theta = I_{\Theta}(x, x+\delta_{t+1})$
\EndIf
\EndFor
\If {$f_x(x+ \delta^*) \leq 0$}
    \State $ub = \min$\{$ub$, $c\}$
    \If {$ub$  $<10^{9}$}
        \State  Update $c \leftarrow$ 
        $(lb + ub)$/2
    \EndIf
\Else
    \State $lb$= $\max \{lb$, $c\}$
    \If {$ub$ $<10^{9}$}
        \State  Update $c \leftarrow$ $(lb + ub)$/2
    \Else
        \State Update $c \leftarrow c \cdot 10$
    \EndIf
\EndIf
\EndFor
\State  {\bfseries Return} $\delta^*$, $I^*_\Theta$
\end{algorithmic}
\end{algorithm}

\subsection{Fixed-Iteration Analysis for Penalty-based Attack}

To compare with the attack success of our MinMax Attack in Table \ref{recons error}, we fix $T$=40 total iterations for penalty-based attack. Specifically, for binary search with $a$ times (each with $b$ iterations), we ran $a \times b =40$ total iterations with ($a$,$b$)=(2,20)/(4,10) on convolutional autoencoder and MNIST. The attack success rate of (2,20)/(4,10)/MinMax(ours) is 9.0/93.54/\textbf{99.86} \%, demonstrating the effectiveness of our attack.

\subsection{Run-time Analysis of Different Data Augmentation Methods}\label{append_runtime}
We perform additional run-time analysis for all methods using the same computing resources on the tasks in Tables \ref{several augs} for MINST and CIFAR-10. Table \ref{tab:run_time_ofAllMethod} shows the average run-time for generating one sample for each method. Note that although MINE-UAE consumes more time than others, its effectiveness in data augmentation is also the most prominent.
\begin{table*}[h]
    \centering
    \begin{tabular}{c|cccc}
     \toprule
    \multicolumn{5}{c}{MNIST}\\
    \toprule
    &MINE-UAE&$L_2$-UAE&GA&Flip or Rotation\\
     \midrule
        avg. per sample &  0.059 sec. &0.025 sec.&	0.012 sec. &$9.18 \times 10^{-5}$ sec.	\\
    \toprule
    \multicolumn{5}{c}{CIFAR-10}\\
    \toprule
    &MINE-UAE&$L_2$-UAE&GA&Flip or Rotation\\
    \midrule
    avg. per sample &	1.735 sec. & 0.37 sec.&	0.029 sec. &0.00022 sec.\\
    \toprule

    \end{tabular}
    \caption{Run-time analysis of different data augmentation methods.}
    \label{tab:run_time_ofAllMethod}
\end{table*}

\newpage

\subsection{Proof of Theorem 1}
\label{subsec_proof}

The proof incorporates the several special structures of this non-convex min-max  problem, e.g., linear coupling between $c$ and $f_x^+(x+\delta)$, the oracle of MINE, etc, which are not explored or considered adequately in prior arts. Therefore, the following theoretical analysis for the proposed algorithm is sufficiently different from the existing works on convergence analysis of min-max algorithms, although some steps/ideas of  deviation are similar, e.g., descent lemma, perturbation term. For the purpose of the completeness, we provide the detailed proof as follows.
To make the analysis more complete, we consider a slightly more general version of Algorithm \ref{minmax}, where in step 7 we use the update rule $c_{t+1}= (1 - \beta \cdot \gamma_t)\cdot c_t + \beta \cdot f_x^+(x+\delta_{t+1})$, where $\gamma_t \geq 0$ and $\gamma_t=1/t^{1/4}$ is a special case used in Algorithm \ref{minmax}. 
%\subsubsection{Proof of Theorem 1}\label{th.pr1}

{\bf \emph{Proof}}: First, we quantity the descent of the objective value after performing one round update of $\delta$ and $c$. Note that the $\delta$-subproblem of the attack formulation is
\begin{equation}
\delta_{t+1}=\arg\min_{\delta: x +\delta \in [0,1]^{d},~\delta \in [-\epsilon,\epsilon]^{d}}\left\langle c_t\nabla f_x^+(x+\delta_t)-I_{\Theta_t}(x, x+\delta_t), \delta-\delta_t\right\rangle+\frac{1}{2\alpha}\|\delta-\delta_t\|^2.
\end{equation}

From the optimality condition of $\delta$, we have
\begin{equation}
    \langle \nabla c_t f_x^+(x+\delta_t)-\nabla I_{\Theta^t}(x,x+\delta_t)+\frac{1}{\alpha}(\delta_{t+1}-\delta_t),\delta_{t+1}-\delta_t\rangle\le 0.\label{eq.optcond}
\end{equation}

According to the gradient Lipschitz continuity of function $f_x^+(x)$ and $I_{\Theta}(x,x+\delta)$, it can be easily checked that $F(\delta,c)$ has gradient
Lipschitz continuity with constant $L_f\bc+L_I$. Then, we are able to have
\begin{align}
    &c_t f_x^+(x+\delta_{t+1})-I_{\Theta_t}(x,x+\delta_{t+1})
    \\
    \le & c_t f_x^+(x+\delta_t) - I_{\Theta_t}(x,x+\delta_t)+\langle c_t\nabla f_x^+(x+\delta_t)+\nabla I_{\Theta_t}(x,x+\delta_t),\delta_{t+1}-\delta_t\rangle +\frac{L_f\bc+L_I}{2}\left\|\delta_{t+1}-\delta_t\right\|^2.\label{eq.lip}
\end{align}

Substituting \eqref{eq.optcond} into \eqref{eq.lip}, we have
\begin{equation}
    c_t f_x^+(x+\delta_{t+1}) - I_{\Theta_t}(x,x+\delta_{t+1})\le c_t f_x^+(x+\delta_{t}) - I_{\Theta_t}(x,x+\delta_{t})-\left(\frac{1}{\alpha}-\frac{L_f\bc+L_I}{2}\right)\left\|\delta_{t+1}-\delta_t\right\|^2.
\end{equation}

From Assumption 2, we can further know that
\begin{equation}
    c_t f_x^+(x+\delta_{t+1}) - I_{\Theta_{t+1}}(x,x+\delta_{t+1})\le c_t f_x^+(x+\delta_{t}) - I_{\Theta_t}(x,x+\delta_{t})-\left(\frac{1}{\alpha}-\frac{L_f\bc+L_I}{2}\right)\left\|\delta_{t+1}-\delta_t\right\|^2+\eta.
\end{equation}
When $\alpha_t\le \frac{1}{L_f\bc+L_I}$, we have
\begin{equation}\label{eq.descentx}
    c_t f_x^+(x+\delta_{t+1}) - I_{\Theta_{t+1}}(x,x+\delta_{t+1})\le c_t f_x^+(x+\delta_{t}) - I_{\Theta_t}(x,x+\delta_{t})-\frac{1}{2\alpha}\left\|\delta_{t+1}-\delta_t\right\|^2+\eta.
\end{equation}

Let function $f'^+_x(c_t,\delta_t)=c_t f_x^+(x+\delta_t)-\mathbbm{1}(c_t)$ and $\xi_t$ denote the subgradient of $\mathbbm{1}(c_t)$, where $\mathbbm{1}(\cdot)$ denotes the indicator function. Since function $cf_x^+(x+\delta)$ is concave with respect to $c$, we have
\begin{align}
\notag
&f'^+_x(c_{t+1},\delta_{t+1})-f'^+_x(c_t, \delta_{t+1})\le \langle f_x^+(x+ \delta_{t+1}), c_{t+1}-c_t\rangle-\langle\xi_t, c_{t+1}-c_t\rangle
\\\notag
=&\langle f_x^+(x+ \delta_{t+1})-  f_x^+(x, \delta_{t}), c_{t+1}-c_t\rangle+\langle f_x^+(x+ \delta_{t}), c_{t+1}-c_t\rangle
\\\notag
&-\langle \xi_{t+1},c_{t+1}-c_t\rangle-\langle\xi_t-\xi_{t+1},c_{t+1}-c_t\rangle
\\\notag
\mathop{=}\limits^{(a)}&\frac{1}{\beta}(c_{t+1}-c_t)^2+\gamma_t c_{t+1} (c_{t+1}-c_t)+(\xi_{t+1}-\xi_t)(c_{t+1}-c_t)
\\\notag
\mathop{=}\limits^{(b)}&\frac{1}{\beta}(c_{t+1}-c_t)^2+\gamma_{t-1} c_{t} (c_{t+1}-c_t)+(f_x^+(x+\delta_{t+1})-f_x^+(x+\delta_t))(c_{t+1}-c_t)
\\\notag
\mathop{\le}\limits^{(c)}&\frac{1}{2\beta}(c_t-c_{t-1})^2+\frac{\beta l^2_f}{2}\|\delta_{t+1}-\delta_t\|^2-(\frac{\gamma_{t-1}}{2}-\frac{1}{\beta})(c_{t+1}-c_t)^2
\\
&+\frac{\gamma_t}{2}c^2_{t+1}-\frac{\gamma_{t-1}}{2}c^2_t+\frac{\gamma_{t-1}-\gamma_t}{2}c^2_{t+1}\label{eq.descenty}
\end{align}
where in $(a)$ we use the optimality condition of $c$-problem, i.e.,
\begin{equation}
\xi_{t+1}-f_x^+(x+\delta_{t+1})+\frac{1}{\beta}(c_{t+1}-c_t)+\gamma_tc_{t+1}=0,
\end{equation}
and in $(b)$ we substitute
\begin{multline}
\langle\xi_{t+1}-\xi_{t}, c_{t+1}-c_t\rangle=\langle f_x^+(x+\delta_{t+1})- f_x^+(x+\delta_t), c_{t+1}-c_t\rangle
\\
-\frac{1}{\beta}\langle \underbrace{c_{t+1}-c_t-(c_t-c_{t-1})}_{:=v_t},c_{t+1}-c_t\rangle -\langle\gamma_t c_{t+1}-\gamma_{t-1} c_t,c_{t+1}-c_t\rangle,
\end{multline}
and in $(c)$ we use the quadrilateral identity  and  according to the Lipschitz continuity function $f_x^+(x+\delta)$, we have
\begin{equation}
\left(f_x^+(x+\delta_{t+1})- f_x^+(x+\delta_t)\right) \left(c_{t+1}- c_t\right)\le\frac{\beta l^2_f}{2}\|\delta_{t+1}-\delta_t\|^2+\frac{1}{2\beta}(c_{t+1}-c_t)^2,
\end{equation}
 and also
\begin{align}
\notag
&\gamma_{t-1}c_t(c_{t+1}-c_t)=\frac{\gamma_{t-1}}{2}\left(c_{t+1}^2-c_t^2-(c_{t+1}-c_t)^2\right)
\\
&=\frac{\gamma_t}{2}c_{t+1}^2-\frac{\gamma_{t-1}}{2}c_t^2-\frac{\gamma_{t-1}}{2}(c_{t+1}-c_t)^2+\left(\frac{\gamma_{t-1}-\gamma_t}{2}\right)c_{t+1}^2.
\end{align}

Combining \eqref{eq.descentx} and  \eqref{eq.descenty}, we have the descent of the objective function, i.e.,
\begin{multline}
c_{t+1}f_x^+(x+\delta_{t+1})+I_{\Theta_{t+1}}(x,x+\delta_{t+1})-\left(c_tf_x^+(x+\delta_t)+I_{\Theta_t}(x,x+\delta_{t})\right)
\\
\le\frac{1}{2\beta}(c_t-c_{t-1})^2-\left(\frac{1}{2\alpha}-\frac{\beta l^2_f}{2}\right)\|\delta_{t+1}-\delta_t\|^2-\left(\frac{\gamma_{t-1}}{2}-\frac{1}{\beta}\right)(c_{t+1}-c_t)^2
\\
+\frac{\gamma_t}{2}c_{t+1}^2-\frac{\gamma_{t-1}}{2}c_t^2+\frac{\gamma_{t-1}-\gamma_t}{2}c_{t+1}^2+\eta.\label{eq.descentlemma}
\end{multline}

Second,  we need to obtain the recurrence of the size of the difference between two consecutive iterates. Note that the maximization problem is
\begin{equation}
c_{t+1}=\arg\max_{0\le c\le \bc} cf_x^+(x+\delta_{t+1})-\frac{1}{2\beta}(c-c_t)^2 -\gamma_t c^2,
\end{equation}
and the update of sequence $\{c_t\}$ can be implemented very efficiently as stated in the algorithm description as
\begin{equation}\label{eq.updc}
    c_{t+1}=P_{\mathcal{C}}\left((1-\beta\gamma_t)c_t+\beta f_x^+(x+\delta_{t+1})\right).
\end{equation}

%Note that $f(x+\delta)$ is upper bounded by $F'$, so sequence $\{c_t,\forall t\ge1\}$ is upper bounded as well, where the upper bound is given below:
%\begin{align}
%    \|c_{t+1}\|\mathop{\le}\limits^{(a)}& \underbrace{|1-\beta\gamma_t|}_{:=\rho_t}\|c_t\|+\beta F'
 %   \\
%    \le&\rho^2_t\|c_{t-1}\|+\rho_t \beta F' + \beta F'
%    \\
%    \mathop{\le}\limits^{(b)} & \frac{\beta F'}{1-\rho_t}:\le\frac{ F'}{\gamma_t}=c^u_t\label{eq.upc}
%\end{align}
%where in $(a)$ we use to the triangular inequality and require $1-\beta\gamma_t>-1$, i.e., $\gamma_t<2/\beta$; and $(b)$ is true due to $c_0:=0$. 

From the optimality condition of $c$-subproblem at the $t+1$th iteration, we have
\begin{equation}\label{eq.yopt1}
-\langle f_x^+(x+\delta_{t+1})-\frac{1}{\beta}(c_{t+1}-c_t)-\gamma_tc_{t+1}, c_{t+1}-c\rangle\le 0, \forall c\in\mathcal{C}
\end{equation}
also, from the optimality condition of $c$-subproblem at the $t$th iteration, we have
\begin{equation}\label{eq.yopt2}
-\langle f_x^+(x+\delta_{t})-\frac{1}{\beta}(c_{t}-c_{t-1})-\gamma_{t-1}c_{t}, c-c_t\rangle\ge 0, \forall c\in\mathcal{C}.
\end{equation}

Plugging in $c=c_t$ in \eqref{eq.yopt1}, $c=c_{t+1}$ in \eqref{eq.yopt2} and combining them together, we can get
\begin{equation}\label{eq.recur}
\frac{1}{\beta}v_{t+1}(c_{t+1}-c_t)+ (\gamma_t c_{t+1}-\gamma_{t-1} c_t)(c_{t+1}-c_t)\le (f_x^+(x+\delta_{t+1})- f_x^+(x+\delta_t))(c_{t+1}-c_t).
\end{equation}

In the following, we will use this above inequality to analyze the recurrence of the size of the difference between two consecutive iterates. Note that
\begin{align}
\notag
&(\gamma_t c_{t+1}-\gamma_{t-1} c_t)(c_{t+1}-c_t)=(\gamma_tc_{t+1}-\gamma_t c_t+\gamma_t c_t-\gamma_{t-1}c_t)(c_{t+1}-c_t)
\\\notag
=&\gamma_t(c_{t+1}-c_t)^2+(\gamma_t-\gamma_{t-1})  c_t(c_{t+1}-c_t)
\\\notag
=&\gamma_t(c_{t+1}-c_t)^2+\frac{\gamma_t-\gamma_{t-1}}{2}\left(c_{t+1}^2-c_t^2-(c_{t+1}-c_t)^2\right),
\\
=&\frac{\gamma_{t}+\gamma_{t-1}}{2}(c_{t+1}-c_t)^2-\frac{\gamma_{t-1}-\gamma_t}{2}\left(c_{t+1}^2-c_t^2\right)\label{eq.gammar}
\end{align}
and
\begin{equation}
v_{t+1}(c_{t+1}-c_t)=\frac{1}{2}\left((c_{t+1}-c_t)^2+v_{t+1}^2-(c_t-c_{t-1})^2\right).\label{eq.eqrel}
\end{equation}

Substituting \eqref{eq.gammar} and \eqref{eq.eqrel} into \eqref{eq.recur}, we have
\begin{align}\notag
&\frac{1}{2\beta}(c_{t+1}-c_t)^2-\frac{\gamma_{t-1}-\gamma_t}{2}c_{t+1}^2
\\
\le & \frac{1}{2\beta}(c_t-c_{t-1})^2-\frac{1}{2\beta}v_{t+1}^2-\frac{\gamma_{t-1}-\gamma_t}{2}c_t^2-\frac{\gamma_{t-1}+\gamma_t}{2}(c_{t+1}-c_t)^2
+(f_x^+(x+\delta_{t+1})- f_x^+(x+\delta_t))(c_{t+1}-c_t)
\\
\mathop{\le}\limits^{(a)}&\frac{1}{2\beta}(c_t-c_{t-1})^2-\gamma_t(c_{t+1}-c_t)^2-\frac{\gamma_{t-1}-\gamma_t}{2}c_t^2+(f_x^+(x+\delta_{t+1})- f_x^+(x+\delta_t))(c_{t+1}-c_t)
\\
\mathop{\le}\limits^{(b)}&\frac{1}{2\beta}(c_t-c_{t-1})^2-\gamma_t(c_{t+1}-c_t)^2-\frac{\gamma_{t-1}-\gamma_t}{2}c_t^2+\frac{l^2_f}{2\gamma_t}\|\delta_{t+1}-\delta_t\|^2+\frac{\gamma_t}{2}(c_{t+1}-c_t)^2
\\
\le&\frac{1}{2\beta}(c_t-c_{t-1})^2-\frac{\gamma_t}{2}(c_{t+1}-c_t)^2-\frac{\gamma_{t-1}-\gamma_t}{2}c_t^2+\frac{l^2_f}{2\gamma_t}\|\delta_{t+1}-\delta_t\|^2
\end{align}
where $(a)$ is true because $0<\gamma_t<\gamma_{t-1}$; in $(b)$ we use Young's inequality.

Multiplying by 4 and dividing by $\beta\gamma_t$ on the both sides of the above equation, we can get
\begin{align}\notag
&\frac{2}{\beta^2\gamma_t}(c_{t+1}-c_t)^2-\frac{2}{\beta}\left(\frac{\gamma_{t-1}}{\gamma_t}-1\right)c_{t+1}^2
\\
\le &\frac{2}{\beta^2\gamma_t}(c_t-c_{t-1})^2-\frac{2}{\beta}\left(\frac{\gamma_{t-1}}{\gamma_t}-1\right)c_{t}^2
-\frac{2}{\beta}(c_{t+1}-c_t)^2+\frac{2l^2_f}{\beta\gamma_t^2}\|\delta_{t+1}-\delta_t\|^2
\\\notag
\le&\frac{2}{\beta^2\gamma_{t-1}}(c_{t}-c_{t-1})^2-\frac{2}{\beta}\left(\frac{\gamma_{t-2}}{\gamma_{t-1}}-1\right)c_{t}^2+\frac{2}{\beta}\left(\frac{1}{\gamma_t}-\frac{1}{\gamma_{t-1}}\right)(c_t-c_{t-1})^2
\\
&+\frac{2}{\beta}\left(\frac{\gamma_{t-2}}{\gamma_{t-1}}-\frac{\gamma_{t-1}}{\gamma_t}\right)c_t^2
-\frac{2}{\beta}(c_{t+1}-c_t)^2+\frac{2l^2_f}{\beta\gamma_t^2}\|\delta_{t+1}-\delta_t\|^2.\label{eq.diffiter}
\end{align}

Combining \eqref{eq.descentlemma} and \eqref{eq.diffiter}, we have
\begin{align}
\notag
&c_{t+1}f_x^+(x+\delta_{t+1})+I_{\Theta_{t+1}}(x,x+\delta_{t+1})-\frac{\gamma_t}{2}c_{t+1}^2+\frac{2}{\beta^2\gamma_t}(c_{t+1}-c_t)^2-\frac{2}{\beta}\left(\frac{\gamma_{t-1}}{\gamma_t}-1\right)c_{t+1}^2
\\\notag
\le & c_tf_x^+(x+\delta_t)+I_{\Theta_t}(x,x+\delta_{t})-\frac{\gamma_{t-1}}{2}c_t^2+\frac{2}{\beta^2\gamma_{t-1}}(c_t-c_{t-1})^2-\frac{2}{\beta}\left(\frac{\gamma_{t-2}}{\gamma_{t-1}}-1\right)c_t^2
\\\notag
&+\frac{1}{2\beta}(c_t-c_{t-1})^2-\frac{1}{\beta}(c_{t+1}-c_t)^2-\left(\frac{1}{2\alpha_t}-\left(\frac{\beta L^2_f}{2}+\frac{2L^2_f}{\beta\gamma_t^2}\right)\right)\|\delta_{t+1}-\delta_t\|^2
\\
&+\frac{\gamma_{t-1}-\gamma_t}{2}c_{t+1}^2+\frac{2}{\beta}\left(\frac{1}{\gamma_t}-\frac{1}{\gamma_{t-1}}\right)(c_{t}-c_{t-1})^2+\frac{2}{\beta}\left(\frac{\gamma_{t-2}}{\gamma_{t-1}}-\frac{\gamma_{t-1}}{\gamma_t}\right)c_t^2+\eta.\label{eq.relatp}
\end{align}

Third, we construct the potential function to measure the descent achieved by the sequence. From \eqref{eq.relatp}, we have
\begin{align}
\notag
&\mathcal{P}_{t+1}\le \mathcal{P}_t-\frac{1}{2\beta}(c_{t+1}-c_t)^2-\left(\frac{1}{2\alpha}-\left(\frac{\beta l^2_f}{2}+\frac{2l^2_f}{\beta\gamma_t^2}\right)\right)\|\delta_{t+1}-\delta_t\|^2
\\
&+\frac{\gamma_{t-1}-\gamma_t}{2}c_{t+1}^2+\frac{2}{\beta}\left(\frac{1}{\gamma_{t+1}}-\frac{1}{\gamma_{t}}\right)(c_{t+1}-c_{t})^2+\frac{2}{\beta}\left(\frac{\gamma_{t-2}}{\gamma_{t-1}}-\frac{\gamma_{t-1}}{\gamma_t}\right)c_t^2+\eta\label{eq.whole}
\end{align}
where we define
\begin{equation}
\mathcal{P}_t:=c_{t}f(x+\delta_{t})+I_{\Theta_{t}}(x,x+\delta_{t})+\left(\frac{1}{2\beta}+\frac{2}{\beta^2\gamma_{t-1}}+\frac{2}{\beta}\left(\frac{1}{\gamma_{t}}-\frac{1}{\gamma_{t-1}}\right)\right)(c_t-c_{t-1})^2-\frac{\gamma_{t-1}}{2}c_t^2-\frac{2}{\beta}\left(\frac{\gamma_{t-2}}{\gamma_{t-1}}-1\right)c_t^2
\end{equation}

To have the descent of the potential function, it is obvious that we need to require 
\begin{equation}
-\frac{1}{2\beta}+\frac{2}{\beta}\left(\frac{1}{\gamma_{t+1}}-\frac{1}{\gamma_t}\right)<0,\label{eq.shrink}
\end{equation}
which is equivalent to condition
$1/\gamma_{t+1}-1/\gamma_t\le 0.25$ so that the sign in the front of term $(c_{t+1}-c_t)^2$ is negative.

When $1/\gamma_{t+1}-1/\gamma_t\le 0.2$, we have
\begin{equation}
\mathcal{P}_{t+1}\le \mathcal{P}_{t} -\left(\frac{1}{2\alpha}-\left(\frac{\beta l^2_f}{2}+\frac{2l^2_f}{\beta\gamma_t^2}\right)\right)\|\delta_{t+1}-\delta_t\|^2-\frac{1}{10\beta}(c_{t+1}-c_t)^2
+\frac{\gamma_{t-1}-\gamma_t}{2}c_{t+1}^2+\frac{2}{\beta}\left(\frac{\gamma_{t-2}}{\gamma_{t-1}}-\frac{\gamma_{t-1}}{\gamma_t}\right)c_t^2+\eta,\label{eq.whole2}
\end{equation}
which can be also rewritten as
\begin{equation}\label{eq.recurf}
\left(\frac{1}{2\alpha}-\left(\frac{\beta l^2_f}{2}+\frac{2l^2_f}{\beta\gamma_t^2}\right)\right)\|\delta_{t+1}-\delta_t\|^2+\frac{1}{10\beta}(c_{t+1}-c_t)^2\le\mathcal{P}_{t}-\mathcal{P}_{t+1}+\frac{\gamma_{t-1}-\gamma_t}{2}c_{t+1}^2+\frac{2}{\beta}\left(\frac{\gamma_{t-2}}{\gamma_{t-1}}-\frac{\gamma_{t-1}}{\gamma_t}\right)c_t^2+\eta.
\end{equation}

Finally, we can provide the convergence rate of the MinMax algorithm as the following.

From the definition that 
\begin{equation}
\label{eq.optgap}
\mathcal{L}(\delta,c):=\left[\begin{array}{c}\delta-P_{\Delta}[\delta-\nabla_{\delta} F(\delta,c)] \nonumber
\\
c-P_{\mathcal{C}}[c+\nabla_{c} F(\delta,c)] \end{array}\right],
\end{equation}
we know that
\begin{align}
\notag
&\|\mathcal{L}(\delta_t, c_t)\|
\\\notag
\le&\|\delta_{t+1}-\delta_t\|+\|\delta_{t+1}-P_{\Delta}(\delta_t-\nabla_{\delta} F(\delta_t,c_t))\|
+|c_{t+1}-c_t|+|c_{t+1}-P_{\mathcal{C}}(c_t+\nabla_c F(\delta_t, c_t))|
\\\notag
\mathop{\le}\limits^{(a)}&\|\delta_{t+1}-\delta_t\|+\left\|P_{\Delta}\left(\delta_{t+1}-(\nabla_{\delta} F(\delta_{t},c_t)+\frac{1}{\alpha_t}(\delta_{t+1}-\delta_t))\right)-P_{\Delta}(\delta_t-\nabla_{\delta} F(\delta_t,c_t))\right\|
\\\notag
&+|c_{t+1}-c_t|+\left|P_{\mathcal{C}}\left(c_{t+1}+ f(\delta^{t+1})-\frac{1}{\beta}(c_{t+1}-c_t)-\gamma_tc_{t+1}\right)-P_{\mathcal{C}}(c_t+ f(\delta_t))\right|
\\\notag
\mathop{\le}\limits^{(b)}&\left(2+\frac{1}{\alpha}\right)\|\delta_{t+1}-\delta_t\|+\left(2+\frac{1}{\beta}\right)|c_{t+1}-c_t|+\gamma_tc_{t+1}
+| f(\delta_{t+1})- f(\delta_{t})|
\\
\mathop{\le}\limits^{(c)}&\left(2+\frac{1}{\alpha}+l_f\right)\|\delta_{t+1}-\delta_t\|+\left(2+\frac{1}{\beta}\right)|c_{t+1}-c_t|+\gamma_t c_{t+1}
\end{align}
where in $(a)$ we the optimality condition of subproblems; in $(b)$  we use the triangle inequality and non-expansiveness of the projection operator; and $(c)$ is true due to the
Lipschitz continuity.

Then, we have
\begin{equation}
\|\mathcal{L}(\delta_t, c_t)\|^2\le 3\left(2+\frac{1}{\alpha}+l_f\right)^2\|\delta_{t+1}-\delta_t\|^2+3\left(2+\frac{1}{\beta}\right)^2(c_{t+1}-c_t)^2+3\gamma^2_t \bc^2.\label{eq.recurf2}
\end{equation}
When $\alpha\sim\gamma^2_t\sim\eta\sim\mathcal{O}(1/\sqrt{T})$ , and 
\begin{equation}
\frac{1}{2\alpha}>\left(\frac{\beta l^2_f}{2}+\frac{2l^2_f}{\beta\gamma_t^2}\right),
\end{equation}
then from \eqref{eq.recurf} we can know that there exist constants $C_1$ and $C_2$ such that
\begin{equation}\label{eq.refin1}
C_1\sqrt{T}\|\delta_{t+1}-\delta_t\|^2+\frac{1}{10\beta}(c_{t+1}-c_t)^2\le\mathcal{P}_{t}-\mathcal{P}_{t+1}+\frac{\gamma_{t-1}-\gamma_t}{2}c_{t+1}^2+\frac{2}{\beta}\left(\frac{\gamma_{t-2}}{\gamma_{t-1}}-\frac{\gamma_{t-1}}{\gamma_t}\right)\bc^2+\frac{C_2}{\sqrt{T}},
\end{equation}
and from \eqref{eq.recurf2}  there exists constant $C_3$ such that
\begin{equation}\label{eq.refin2}
\|\mathcal{L}(\delta_t, c_t)\|^2\le C_3T\|\delta_{t+1}-\delta_t\|^2+3\left(2+\frac{1}{\beta}\right)^2(c_{t+1}-c_t)^2+3\gamma^2_t \bc^2
\end{equation}

Combining \eqref{eq.refin1} and \eqref{eq.refin2}, we have
\begin{multline}
\|\mathcal{L}(\delta_t, c_t)\|^2
\\
\le\max\left\{\max\{C_1,C_3\}\sqrt{T}, \max\left\{\frac{\sqrt{T}}{10\beta},3\left(2+\frac{1}{\beta}\right)^2\right\}\right\}\left(\mathcal{P}_{t}-\mathcal{P}_{t+1}+\frac{\gamma_{t-1}-\gamma_t}{2}c_{t+1}^2+\frac{2}{\beta}\left(\frac{\gamma_{t-2}}{\gamma_{t-1}}-\frac{\gamma_{t-1}}{\gamma_t}\right)\bc^2\right)
\\
+3\gamma^2_t \bc^2+\frac{C_2}{\sqrt{T}}.
\end{multline}
Hence, there exist constant $C_4$ such that
\begin{equation}
\|\mathcal{L}(\delta_t, c_t)\|^2
\le\sqrt{T}C_4\left(\mathcal{P}_{t}-\mathcal{P}_{t+1}+\frac{\gamma_{t-1}-\gamma_t}{2}\bc^2+\frac{2}{\beta}\left(\frac{\gamma_{t-2}}{\gamma_{t-1}}-\frac{\gamma_{t-1}}{\gamma_t}\right)\bc^2\right)
+3\gamma^2_t \bc^2+\frac{C_2}{\sqrt{T}}.
\end{equation}

Applying the telescoping sum, we have
\begin{equation}
\frac{1}{T}\sum^T_{t=1}\|\mathcal{L}(\delta_t, c_t)\|^2\le \frac{C_4}{\sqrt{T}}\left((\mathcal{P}_1-\mathcal{P}_{T+1})+\frac{\gamma_0\bc^2}{2}+\frac{2\gamma_{0}\bc^2}{\gamma_{1}\beta}\right)+3\frac{\bc^2}{\sqrt{T}}+\frac{C_2}{\sqrt{T}}.
\end{equation}
According to the definition of $T(\varepsilon')$, we can conclude that there exists constant $C$ such that
\begin{equation}
\|\mathcal{L}(\delta_{T(\varepsilon')}, c_{T(\varepsilon')})\|^2\le\frac{1}{T(\varepsilon')}\sum^{T(\varepsilon')}_{t=1}\|\mathcal{L}(\delta_t, c_t)\|^2\le \frac{C}{\sqrt{T(\varepsilon')}}\sim\mathcal{O}\left(\frac{1}{\sqrt{T(\varepsilon')}}\right),
\end{equation}
which completes the proof. $\hfill\blacksquare$
%\clearpage
%\newpage
%\subsubsection{Proof of Corollary 1}\label{th.pr2}

%{\bf \emph{Proof}}: 
%From Theorem 1 we know that when $\epsilon\to 0$, then $T(\epsilon)\to\infty$. Taking the limit, we can obtain
%\begin{equation}
%    \delta_{t+1}-\delta_t\to 0, \quad\textrm{and}\quad c_{t+1}-c_{t}\to 0.
%\end{equation}
%From \eqref{eq.updc} and the choice of $\gamma_t\sim\mathcal{O}(1/T^{1/4})$, we have 
%\begin{equation}
%    c^*=P_{\mathcal{C}}\left(c^*+\beta f(x+\delta^*)\right).
%\end{equation}
%Since $c^*,f(x+\delta^*),\beta\ge0$, we have
%\begin{equation}
%    c^*=c^*+\beta f(x+\delta^*),
%\end{equation}
%which implies that 
%$$f(x+\delta)\to 0.$$
%Also, from Theorem 1 we have
%\begin{equation}
%    \delta^*=P_{\Delta}(\delta^*-\nabla_{\delta}F(\delta^*,c^*)).
%\end{equation}
%According to \cite[Proposition 3]{lu2019snap}, we %have
%\begin{equation}
%    \langle \nabla_\delta F(\delta^*,c^*),\delta-\delta^*\rangle\ge 0,\forall \delta\in\Delta,
%\end{equation}
%which completes the proof. 
%$\hfill\blacksquare$

\clearpage

\subsection{More Discussion on UAEs and On-Manifold Adversarial Examples}

Following \cite{stutz2019disentangling}, adversarial examples can be
either on or off the data manifold, and only the on-manifold
data samples are useful for model generalization. In the supervised setting, without the constraint on finding on-manifold examples, it is shown that simply using norm-bounded adversarial examples for adversarial training will hurt model generalization\footnote{\url{https://davidstutz.de/on-manifold-adversarial-training-for-boosting-generalization}} \cite{stutz2019disentangling}.

As illustrated in Table \ref{tab:UAE_illustration}, in the unsupervised setting we propose to use
the training loss constraint to find on-manifold adversarial
examples and show they can improve model performance.

\subsection{More Details and Results on Data Augmentation and Model Retraining}
\label{appen_training}
%We demonstrate that UAEs improve representation learning in Section 4.4. 

The attack success rate (ASR) measures the effectiveness of finding UAEs. In Table \ref{recons error}, sparse autoencoder has low ASR due to feature selection, making attacks more difficult to succeed. MNIST's ASR can be low because its low reconstruction loss limits the feasible space of UAEs.

Table \ref{ae_detail} summarizes the training epochs and training loss of the models and datasets used in Section \ref{subsec_reconst}. 
\begin{table*}[h]
  \centering
  \begin{adjustbox}{max width=\textwidth}
  \begin{tabular}{c|cccc|ccccc}
    \toprule
     \multicolumn{10}{c}{MNIST} \\
     \midrule
      & \multicolumn{4}{c}{Training Epochs}&  \multicolumn{5}{|c}{Reconstruction Error (training set) }\\
     \midrule
         Autoencoder & Original & \makecell[c]{MINE-UAE} & $L_{2}$-UAE & \makecell[c]{GA\\ ($\sigma=10^{-2}/10^{-3}$)} &Original& \makecell[c]{MINE-UAE } &   $L_{2}$-UAE & \makecell[c]{GA\\ ($\sigma=10^{-2}$)}&\makecell[c]{GA\\ ($\sigma=10^{-3}$)}   \\
    \midrule
     Sparse        & 50 & 80 & 80 & 80 & 0.00563 & 0.00233 &0.00345&0.00267$\pm$2.93e-05&0.00265$\pm$3.60e-5\\ 
     Dense         & 20 & 30 & 30 & 30 & 0.00249 & 0.00218 & 0.00275 &0.00231$\pm$0.00013& 0.0023$\pm$0.00011\\
     \makecell[c]{Convolutional} & 20 & 30& 30& 30& 0.00301& 0.00260& 0.00371& 0.00309$\pm$0.00013& 0.00310$\pm$0.00015\\
     \makecell[c]{Adversarial}   & 50 & 80& 80& 80& 0.044762 & 0.04612& 0.06063 &0.058711$\pm$0.00659 & 0.05551$\pm$0.00642\\
     %\midrule
     \toprule
     \multicolumn{10}{c}{SVHN} \\
     \midrule
     Sparse         & 50 & 80 & 80 & 80 &0.00729 &0.00221 &0.00290 &0.00283$\pm$0.00150 & 
      0.00275 $\pm$ 0.00081\\
     Dense          & 30 & 50 & 50 & 50 &0.00585 &0.00419 & 0.00503 & 0.00781$\pm$0.00223 & 0.00781$\pm$0.00187\\
     \makecell[c]{Convolutional} & \multicolumn{4}{c|}{\makecell[c]{We set 100 epochs, but the training loss converges  \\after 5 epochs}}&0.00140 & 0.00104 & 0.00131&0.00108$\pm$3.83e-05& 0.00113$\pm$6.76e-05\\
     \makecell[c]{Adversarial}  & \multicolumn{4}{c|}{\makecell[c]{We set 200 epochs and use the\\ model with the lowest training loss}}&0.00169 & 0.00124 & 0.02729 & 0.00158$\pm$0.00059 &0.00130$\pm$0.00036\\
     %\midrule
    \bottomrule
\end{tabular}
\end{adjustbox}
\caption{Details on the training epochs and losses for the datasets and models used in Section \ref{subsec_reconst}. The reconstruction error is the average $L_2$ reconstruction loss of the training set.}  
\label{ae_detail}
\end{table*}

In Table \ref{concrete}, concrete autoencoder on Coil-20 has low ASR and the performance does not improve when using MINE-UAE for data augmentation. The original training loss is 0.00683 while the MINE-UAE training loss is increased to 0.01368. With careful parameter tuning, we were still unable to improve the MINE-UAE training loss. Therefore, we conclude that the degraded performance in Coil-20 after data augmentation is likely due to the imitation of feature selection protocol and the model learning capacity.

\subsection{CIFAR-10 and Tiny-ImageNet Results}
\label{appen_recon_large}

Table \ref{cifar-tiny}  shows the results using convolutional autoencoders on CIFAR-10 and Tiny-ImageNet using the same setup as in Table \ref{recons error}. Similar to the findings in Table \ref{recons error} based on MNIST and SVHN, MINE-UAE attains the best loss compared to $L_2$-UAE and GA.

\begin{table}[h]
  \centering
  \begin{adjustbox}{max width=1\textwidth}
  \begin{tabular}{ccc}
    \toprule
     \multicolumn{3}{c}{CIFAR-10 (original loss: 0.00669)~/~\textcolor{brown}{Tiny-ImageNet (original loss: 0.00979)}} \\
     \midrule
      \multicolumn{3}{c}{Reconstruction Error (test set)~/~ASR (training set)}  \\
     \midrule
     \makecell[c]{MINE-UAE} & $L_{2}$-UAE &\makecell[c]{GA ($\sigma=10^{-3}$)}\\
     \midrule
      \makecell[c]{\textbf{0.00562} ($\textcolor{ForestGreen}{\uparrow15.99\%}$) / 99.98\%} &\makecell[c]{0.00613 ($\textcolor{ForestGreen}{\uparrow8.37\%}$) / 89.84\%} &\makecell[c]{0.00564$\pm$1.31e-04  ($\textcolor{ForestGreen}{\uparrow15.70\%}$) / 99.90\% }\\
      \makecell[c]{\textcolor{brown}{\textbf{0.00958}} ($\textcolor{ForestGreen}{\uparrow 21.45\%}$)}/ \textcolor{brown}{99.47\%}&\makecell[c]{\textcolor{brown}{0.00969} ($\textcolor{ForestGreen}{\uparrow1.02\%}$) / \textcolor{brown}{88.02\%}}& \textcolor{brown}{0.00963$\pm$5.31e-05} ($\textcolor{ForestGreen}{\uparrow1.63\%}$) / \textcolor{brown}{99.83\%}\\
     \toprule
     %\multicolumn{7}{c}{Tiny-ImageNet} %\\
     
  \end{tabular}
\end{adjustbox}
\caption{Comparison of data reconstruction by retraining the convolutional autoencoder on UAE-augmented data using CIFAR-10 (first row) and Tiny-ImageNet (second row). The reconstruction error is the average $L_2$ reconstruction loss of the test set. The improvement (in \textcolor{ForestGreen}{green}/\textcolor{red}{red}) is relative to the original model. The 
  attack success rate (ASR) is the fraction of augmented training data having smaller reconstruction loss than the original loss (see Table \ref{tab:UAE_illustration} for definition).   }
  \label{cifar-tiny}
\end{table}

\subsection{More Details on Table \ref{several augs}}
\label{appen_add_aug}

For all convolutional autoencoders, we use 100 epochs and early stopping if training loss converges at early stage.
For data augmentation of the training data, we set the $\text{rotation angle}=10$ and use both horizontal and vertical flip. For Gaussian noise, we use zero-mean and set $\sigma=0.001$. 

%\subsection{Run-Time Analysis}
% Although our computational cost will be higher than Gaussian augmentation (GA),
%\label{subsec_run_time} 
 
%Table \ref{several augs} shows our approach is complementary to GA and can further boost model performance. 
%For unsupervised tasks, the average wall clock time for generating one UAE on MNIST dataset is 0.059 second.

\subsection{Hyperparameter Sensitivity Analysis}
\label{appen_sensitivity}

All of the aforementioned experiments were using $\beta=0.1$ and $\alpha=0.01$. Here we show the results on MNIST with convolution autoencoder (Section \ref{subsec_reconst}) and
the concrete autoencoder (Section \ref{subsec_representation}) using different combinations of $\alpha$ and $\beta$ values. The results are comparable, suggesting that our MINE-based data augmentation is robust to a wide range of hyperpameter values.

\begin{table*}[h]
    \centering
    \begin{adjustbox}{max width=\textwidth}
    \begin{tabular}{c|ccc|ccc}
        \toprule
        \multicolumn{7}{c}{MNIST} \\
        \midrule
        &\multicolumn{3}{c}{Convolution AE}&\multicolumn{3}{|c}{Concrete AE}  \\
        \midrule
        \diagbox {$\alpha$}{$\beta$}&0.05&0.1&0.5&0.05&0.1&0.5\\
        \midrule
        0.01 & 0.00330 & 0.00256 & 0.00283 & 0.01126 & 0.01142 & 0.01134\\
        0.05 & 0.00296 & 0.00278 & 0.00285 & 0.01129 & 0.01133 & 0.01138\\

        \toprule
         
    \end{tabular}
    \end{adjustbox}
    \caption{Comparison of reconstruction error (test set) of convolution and concrete autoencoders with different combinations of $\alpha$ and $\beta$ values on MNIST.}  
\label{tab:hyperparameter}
\end{table*}

\subsection{Ablation Study on UAE's Similarity Objective Function}

As an ablation study to demonstrate the importance of using MINE for crafting UAEs, 
we replace the mutual information (MI) with $L_2$-norm and cosine distance between $conv(x)$ and $conv(x+\delta)$ in our MinMAx attack algorithm.
Table \ref{tab:other_metric} compares the performance of retraining convolutional autoencoders with UAEs generated by the $L_2$-norm , cosine distances and feature scattering-based proposed by \citet{zhang2019defense}. It is observed that using MINE is more effective that these two distances. 

\begin{table}[h]
    \centering
    \begin{adjustbox}{max width=1\textwidth}
    \begin{tabular}{c|c|c}
    \toprule
    \multicolumn{3}{c}{ MNIST (original reconstruction error: 0.00294)}\\
    \midrule
     &  \multicolumn{1}{c}{Reconstruction Error (test set)}&  \multicolumn{1}{|c}{ASR (training set)}\\
    \midrule
    MINE-UAE &\textbf{0.00256} (\textcolor{ForestGreen}{$\uparrow 12.9\%$})& 99.86\%\\
    $L_2$ / cosine    &  \makecell[c]{0.002983 (\textcolor{red}{$\downarrow 14.6\%$}) / 0.002950  (\textcolor{red}{$\downarrow 0.3\%$})}& 96.5\% / 92\%\\
    Feature scattering-based & 0.00268 (\textcolor{ForestGreen}{$\uparrow 8.9\%$)}&99.96\%\\
    \toprule
    \multicolumn{3}{c}{SVHN (original reconstruction error: 0.00128)}\\
    \midrule
    MINE-UAE & \textbf{0.00095} (\textcolor{ForestGreen}{$\uparrow 25.8\%$})&100\%\\
    $L_2$ / cosine & 0.00096  ($\textcolor{ForestGreen}{\uparrow18.75\%}$) / 0.00100 ($\textcolor{ForestGreen}{\uparrow21.88\%}$)&94.52\% / 96.79\% \\
    Feature scattering-based & 0.001 (\textcolor{ForestGreen}{$\uparrow 21.6\%$}) & 99.76\% \\
    \toprule
    
    \end{tabular}
    \end{adjustbox}
       \caption{Ablation Study on UAE's similarity objective function using MinMax algorithm.  The reconstruction error is the average $L_2$ reconstruction loss of the test set. The improvement (in \textcolor{ForestGreen}{green}/\textcolor{red}{red}) is relative to the original model. }
    %\vspace{-4mm}
    %\caption{Data reconstruction following Table 4. The original error is 0.00294 for MNIST and 0.00128 for SVHN.  }
    \label{tab:other_metric}
\end{table}

\subsection{More Data Augmentation Runs}
While the first fun of UAE-based data augmentation is shown to improve model performance, here we explore the utility of more data augmentation runs. 
We conducted two data augmentation runs for sparse autoencoder on SVHN. We re-train the model with $1^{\text{st}}$-run UAEs, $2^{\text{nd}}$-run UAEs (generated by $1^{\text{st}}$-run augmented model) and original training data. 
%The proportion of UAEs and normal training data is $1:1$. 
The reconstruction error on the test set of $2^{\text{nd}}$ data augmentation is 0.00199, which slightly improves the $1^{\text{st}}$-run result (0.00235). In general, we find that $1^{\text{st}}$-run UAE data augmentation has a much more significant performance gain comparing to the $1^{\text{st}}$-run results.

%{\small
%\bibliographystyle{ieee_fullname}
%\bibliography{egbib}
%}

%\clearpage
\subsection{MINE-based Supervised Adversarial Examples for Adversarially Robust Models}
\label{appen_visual_robust}

\begin{figure}[h]
    \centering
    \subfigure[]{\includegraphics[scale=0.14]{ori1.pdf}}
     \subfigure[Madry model (our attack)]{
    \includegraphics[scale=0.22]{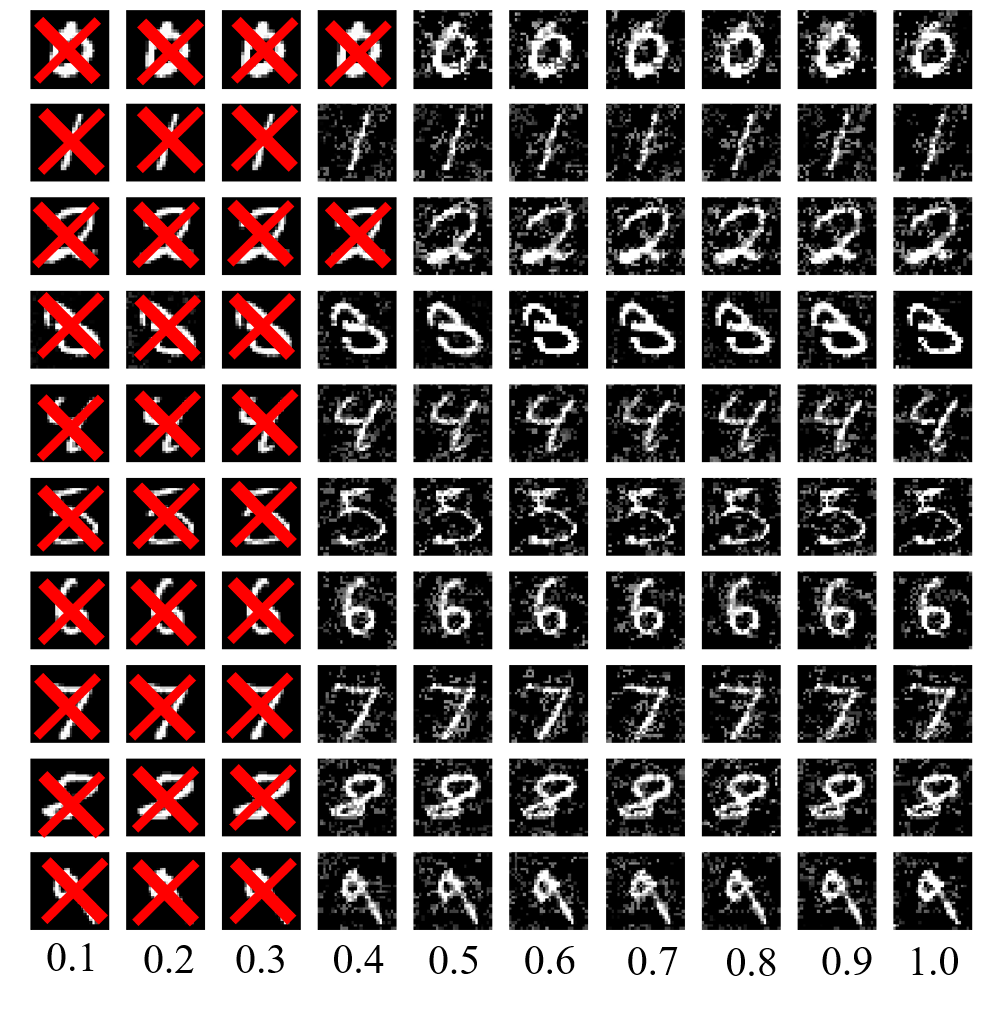}}
    \subfigure[Madry model (PGD)]{
    \includegraphics[scale=0.22]{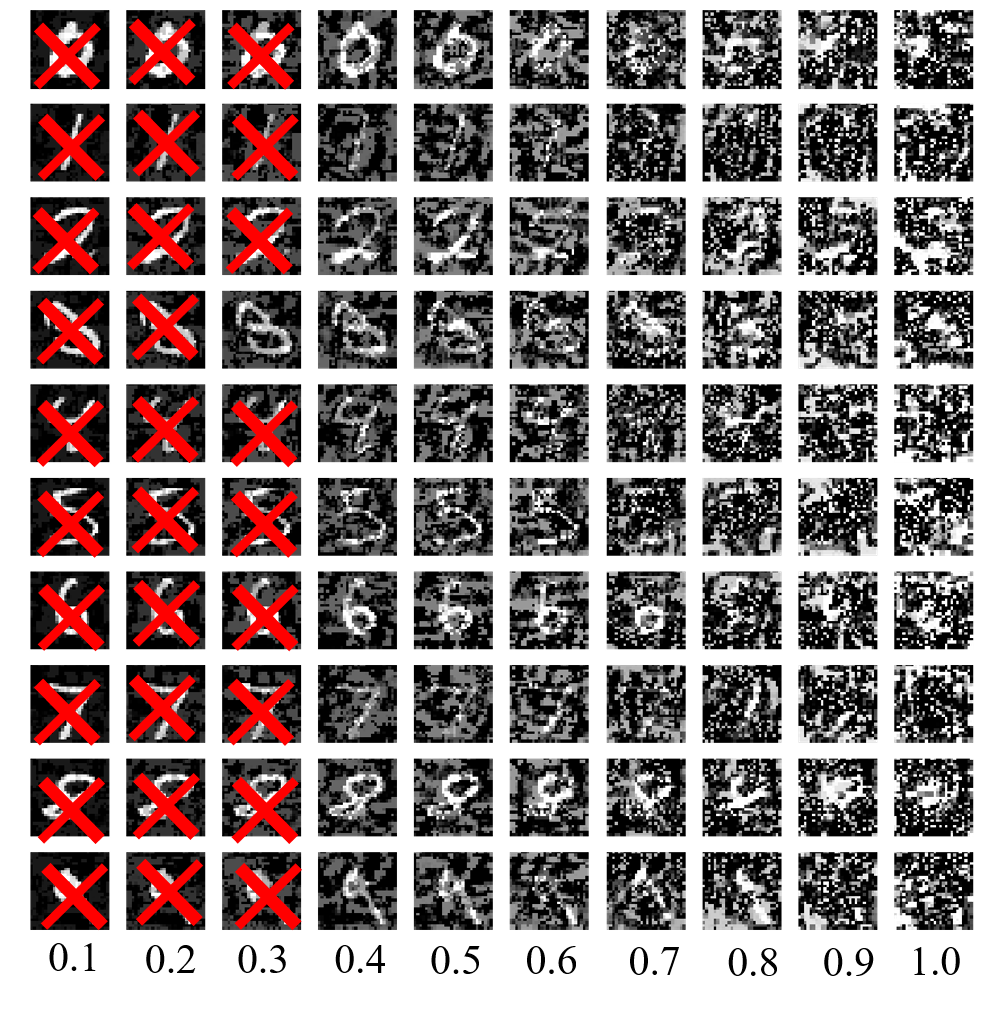}}
    \subfigure[TRADES (our attack)]{
    \includegraphics[scale=0.22]{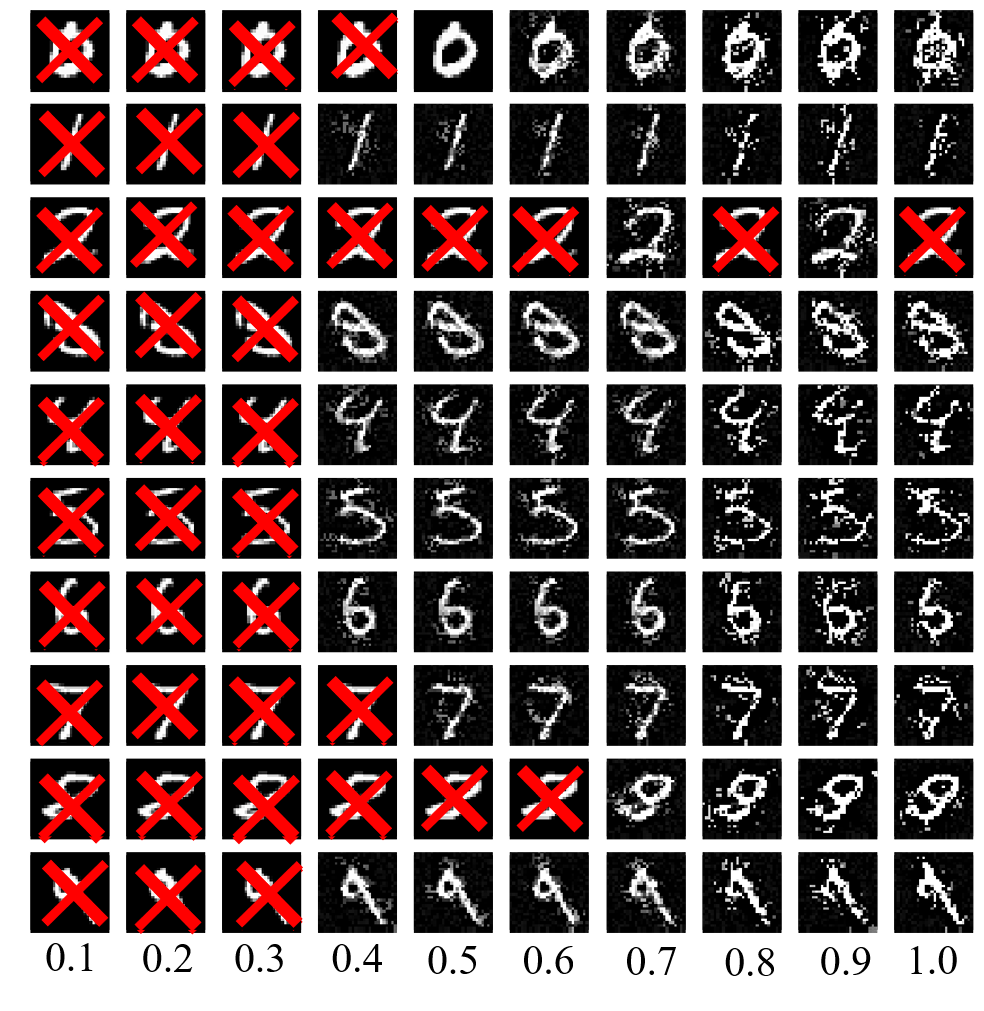}}
    \subfigure[TRADES (PGD)]{
    \includegraphics[scale=0.22]{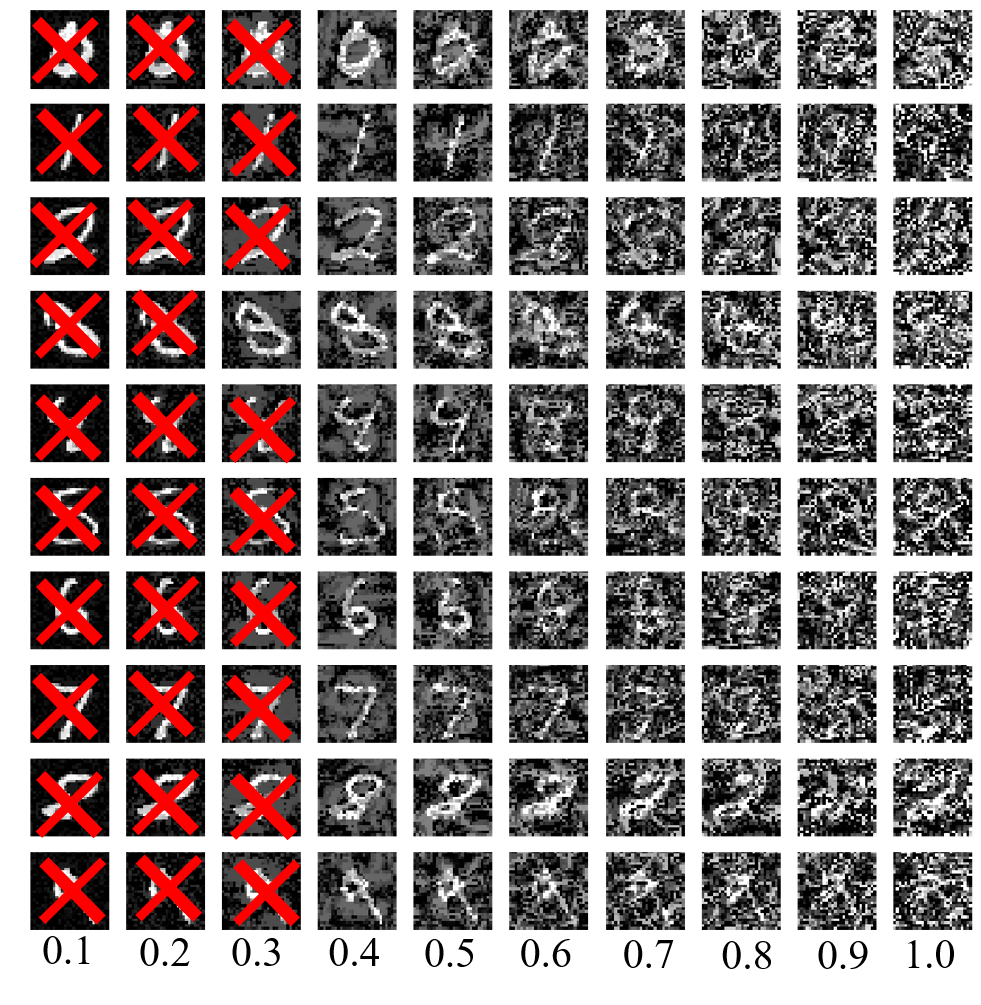}}
    \caption{Visual comparison of untargeted supervised adversarial examples on two released adversarial robust MNIST models: Madry model \cite{madry2017towards} and TRADES \cite{zhang2019theoretically}. We mark the unsuccessful adversarial examples with the red cross. Each row shows the crafted adversarial examples of an original sample. Each column corresponds to different $\epsilon$ values ($L_\infty$-norm perturbation bound) ranging from $0.1$ to $1.0$. }
    \label{Fig_visual_adv_robust}
\end{figure}

\textbf{Visual Comparison of Supervised Adversarial Examples for Adversarially Robust MNIST Models}
Figure \ref{Fig_visual_adv_robust} shows adversarial examples crafted by our attack against the released adversarially robust models trained using the Madry model \cite{madry2017towards} and TRADES \cite{zhang2019theoretically}. Similar to the conclusion in Figure \ref{Fig_visual_MNIST}, our MINE-based attack can generate high-similarity and high-quality adversarial examples for large $\epsilon$ values, while PGD attack fail to do.

\begin{figure}[h]
    \centering
    \begin{minipage}[t]{0.45\linewidth} 
    \centering
     \subfigure[MNIST]{
    \includegraphics[scale=0.25]{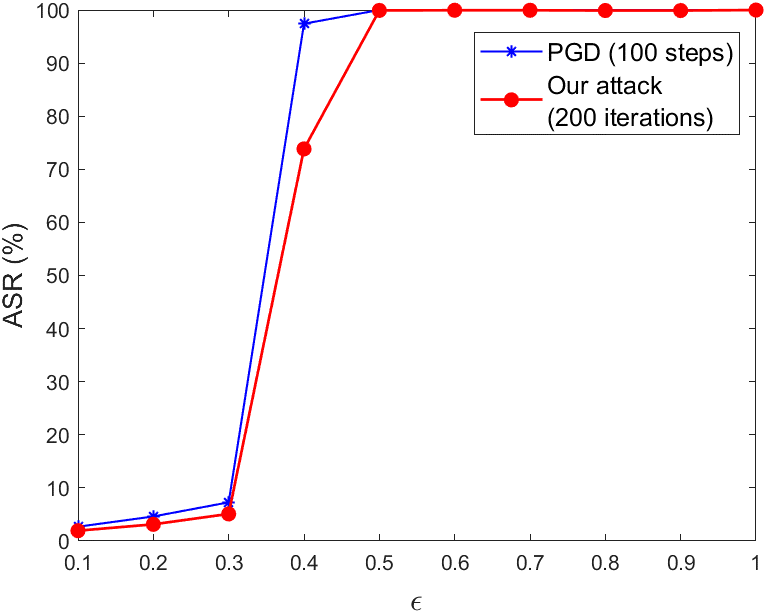}}
    \subfigure[CIFAR-10]{
    \includegraphics[scale=0.25]{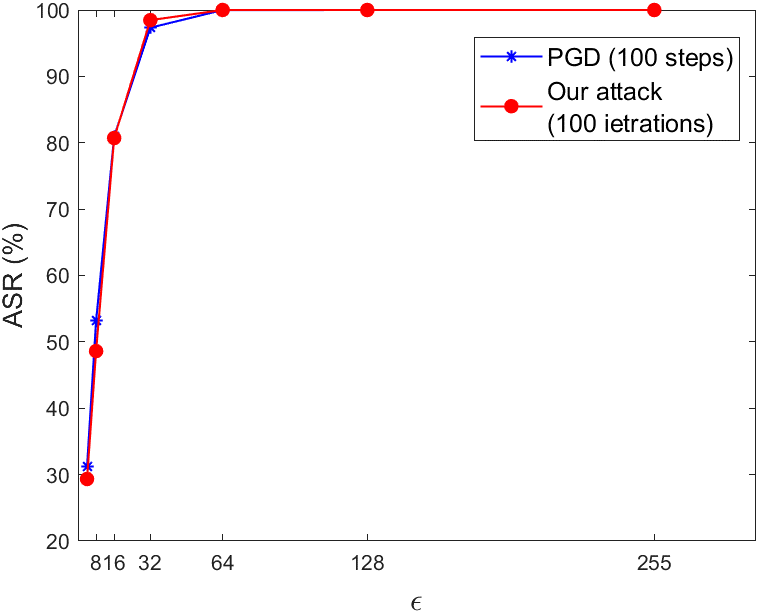}}
    \caption{Attack success rate (ASR) of the released Madry models \cite{madry2017towards} on MNIST and CIFAR-10. }
    \label{Fig_attack_performance_madry}
    \end{minipage}
    \hspace{5mm}
    \begin{minipage}[t]{0.45\linewidth} 
    \centering
    \subfigure[MNIST]{
    \includegraphics[scale=0.25]{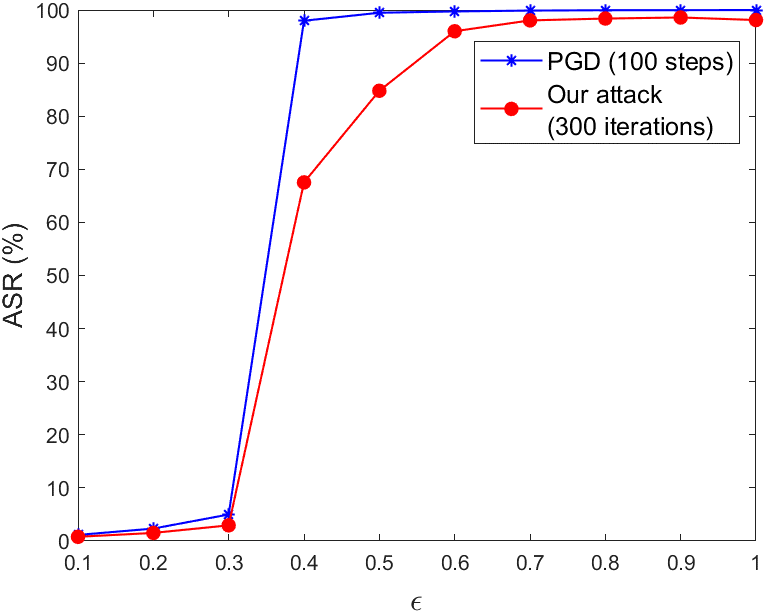}}
    \subfigure[CIFAR-10]{
    \includegraphics[scale=0.25]{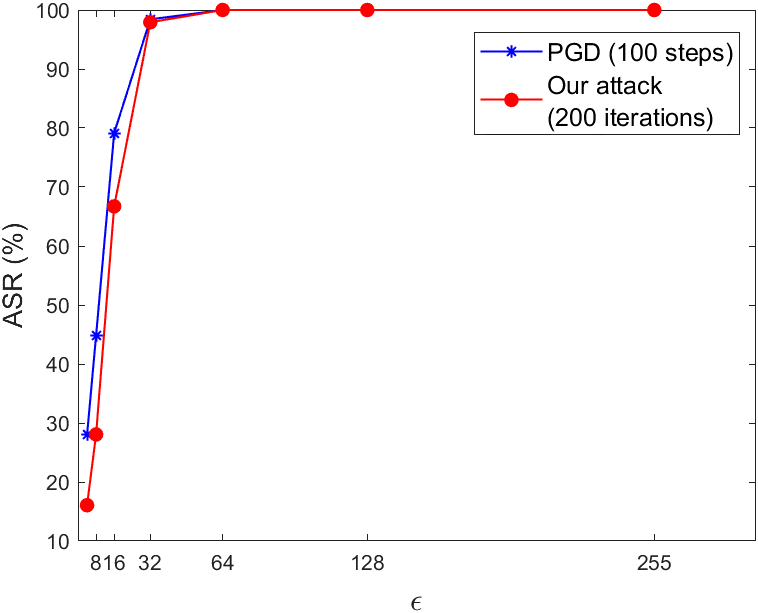}}
    \caption{Attack success rate (ASR) of the released TRADES models on MNIST and CIFAR-10. }
    \label{Fig_attack_performance_trades}
    
\end{minipage}
\end{figure}

\textbf{Attack Performance}
Figure \ref{Fig_attack_performance_madry} and \ref{Fig_attack_performance_trades} show the attack success rate (ASR) of released Madry and TRADES models against our attack and PGD attack on MNIST and CIFAR-10 with different $L_\infty$ threshold $\epsilon$. For all PGD attacks, we use 100 steps and set $\text{step size} = 2.5 * \frac{\epsilon}{\text{number of steps}}$. For all our attacks against Madry model, we use $\text{learning rate} = 0.01$ and run 200 steps for MNIST, 100 steps for CIFAR-10. For all our attacks against TRADES model, we use $\text{learning rate} = 0.01$ and run 300 steps for MNIST, 200 steps for CIFAR-10. Both attacks have comparable attack performance. On MNIST, in the mid-range of the $\epsilon$ values, our attack ASR is observed to be lower than PGD, but it can generate higher-quality adversarial examples as shown in Figure \ref{Fig_visual_adv_robust}.

\begin{figure}[h]
    \centering
     \subfigure[MNIST]{
    \includegraphics[scale=0.3]{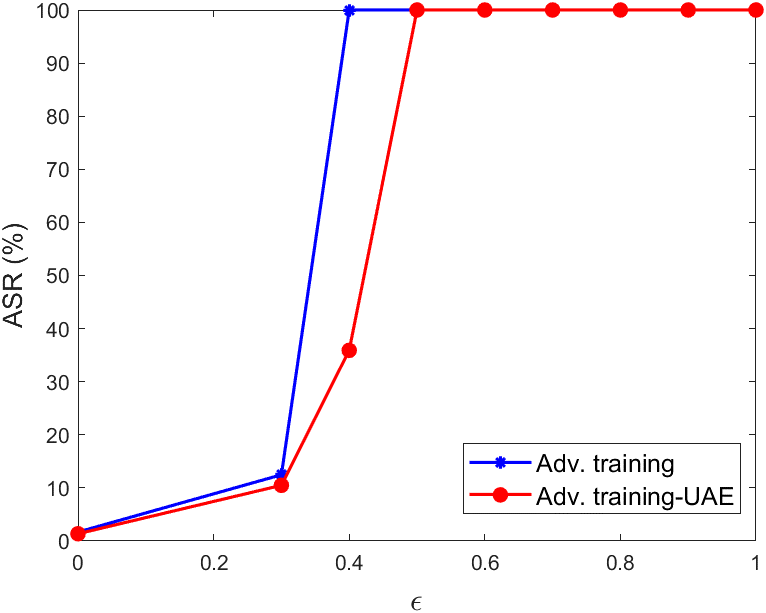}}
    \subfigure[CIFAR-10]{
    \includegraphics[scale=0.3]{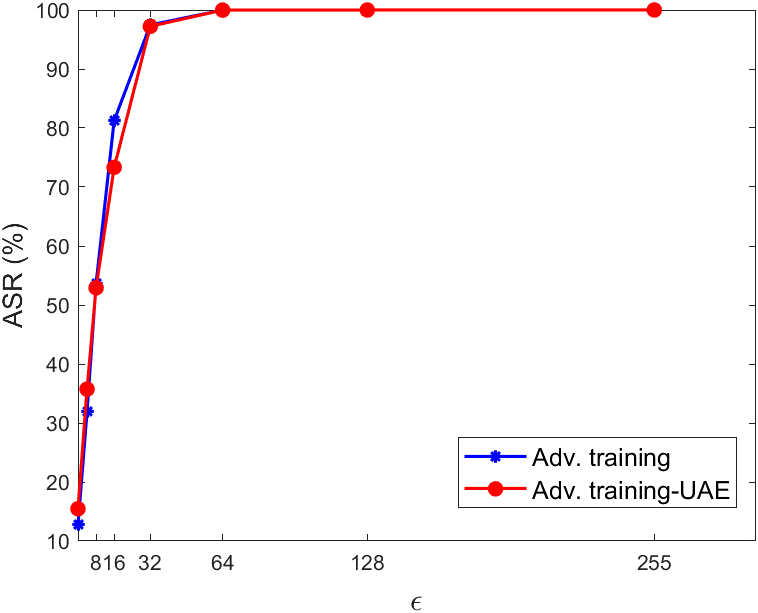}}
    \caption{Performance evaluation of ASR for adv. training and adv. training-UAE against PGD attack with different $\epsilon$ value. Adv. training-UAE consistently shows lower or comparable ASR than adv. training, suggesting that data augmentation using UAE can improve 
adversarial training.}
    \label{Fig_attack_performance_uae}
\end{figure}

\textbf{Adversarial Training with MINE-based Unsupervised Adversarial Examples}
We use the MNIST and CIFAR-10 models in Section \ref{subsec_quali} to compare the performances of standalone adversarial training (Adv. training) \cite{madry2017towards} and adversarial training plus data augmentation by MINE-based unsupervised adversarial examples (Adv. training-UAE) generated from convolutional Autoencoder.  Figure \ref{Fig_attack_performance_uae} shows the attack success rate (ASR) of Adv. training model and Adv training-UAE against PGD attack.  For all PGD attacks, we use 100 steps and set $\text{step size} = 2.5 * \frac{\epsilon}{\text{number of steps}}$. When $\epsilon = 0.4$ , Adv. training-UAE model can still resist more than 60\% of adversarial examples on MNIST. By contrast, ASR is 100\% for Adv. training model. For CIFAR-10, ASR of Adv. training-UAE model is about 8\% lower than Adv. training model when $\epsilon=16$. We therefore conclude that data augmentation using UAE can improve 
adversarial training.

\subsection{Improved Adversarial Robustness after Data Augmentation with MINE-UAEs}
\label{subsec_robustness}
To evaluate the adversarial robustness after data augmentation with  MINE-UAEs,
we use the MNIST and CIFAR-10 models in Section \ref{subsec_quali} and Section \ref{subsec_contrastive}, respectively.
We randomly select 1000 classified correctly images (test set) to generate adversarial examples. For all PGD attacks, We set step size$=0.01$ and use 100 steps.

\begin{figure}[h]
    \centering
     \subfigure[MNIST (convolutional-autoencoder)]{
    \includegraphics[scale=0.48]{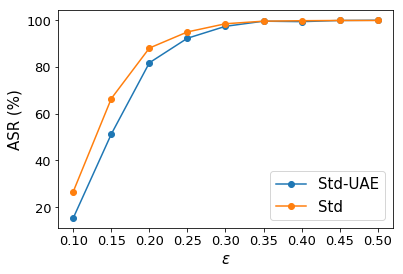}}
    \subfigure[CIFAR-10 (SimCLR)]{
    \includegraphics[scale=0.48]{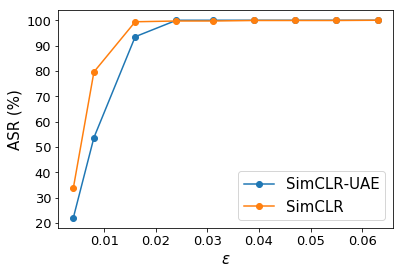}}
    \caption{Robustness evaluation of attack success rate (ASR) for the original model (Std/SimCLR) and the models trained with UAE augmentation (Std-UAE/ SimCLR-UAE) against PGD attack with different $\epsilon$ values. Models trained with UAE shows better robustness (lower ASR) than original models, implying that data augmentation with UAE can strengthen the adversarial robustness.}
    \label{Fig_robustness_performance_uae}
\end{figure}

In our first experiment (Figure \ref{Fig_robustness_performance_uae} (a)),
we train the convolutional classifier (Std) and Std with UAE (Std-UAE) generated from the convolutional autoencoder on MNIST.  The attack success rate (ASR) of the Std-UAE is consistently lower than the Std for each $\epsilon$ value. 

In the second experiment (Figure \ref{Fig_robustness_performance_uae} (b)), the ASR of SimCLR-UAE is significantly lower than that of the original SimCLR model, especially for $\epsilon\leq 0.02$.
When $\epsilon=0.01$, SimCLR-UAE on CIFAR-10 can still resist more than 40\% of adversarial examples, which is significantly better than the original  SimCLR model. 
Based on the empirical results, we therefore conclude that data augmentation using UAE can improve adversarial robustness of unsupervised machine learning tasks.

\end{document}